\PassOptionsToPackage{unicode}{hyperref}
\PassOptionsToPackage{hyphens}{url}
\PassOptionsToPackage{dvipsnames,svgnames,x11names}{xcolor}
\documentclass[
  11pt,
]{article}
\usepackage{xcolor}
\usepackage[margin=1in]{geometry}
\usepackage{amsmath,amssymb}
\usepackage{iftex}
\ifPDFTeX
  \usepackage[T1]{fontenc}
  \usepackage[utf8]{inputenc}
  \usepackage{textcomp} % provide euro and other symbols
\else % if luatex or xetex
  \usepackage{unicode-math} % this also loads fontspec
  \defaultfontfeatures{Scale=MatchLowercase}
  \defaultfontfeatures[\rmfamily]{Ligatures=TeX,Scale=1}
\fi
\usepackage{lmodern}
\ifPDFTeX\else
\fi
\IfFileExists{upquote.sty}{\usepackage{upquote}}{}
\IfFileExists{microtype.sty}{% use microtype if available
  \usepackage[]{microtype}
  \UseMicrotypeSet[protrusion]{basicmath} % disable protrusion for tt fonts
}{}
\usepackage{setspace}
\makeatletter
\@ifundefined{KOMAClassName}{% if non-KOMA class
  \IfFileExists{parskip.sty}{%
    \usepackage{parskip}
  }{% else
    \setlength{\parindent}{0pt}
    \setlength{\parskip}{6pt plus 2pt minus 1pt}}
}{% if KOMA class
  \KOMAoptions{parskip=half}}
\makeatother
\makeatletter
\ifx\paragraph\undefined\else
  \let\oldparagraph\paragraph
  \renewcommand{\paragraph}{
    \@ifstar
      \xxxParagraphStar
      \xxxParagraphNoStar
  }
  \newcommand{\xxxParagraphStar}[1]{\oldparagraph*{#1}\mbox{}}
  \newcommand{\xxxParagraphNoStar}[1]{\oldparagraph{#1}\mbox{}}
\fi
\ifx\subparagraph\undefined\else
  \let\oldsubparagraph\subparagraph
  \renewcommand{\subparagraph}{
    \@ifstar
      \xxxSubParagraphStar
      \xxxSubParagraphNoStar
  }
  \newcommand{\xxxSubParagraphStar}[1]{\oldsubparagraph*{#1}\mbox{}}
  \newcommand{\xxxSubParagraphNoStar}[1]{\oldsubparagraph{#1}\mbox{}}
\fi
\makeatother

\usepackage{longtable,booktabs,array}
\usepackage{calc} % for calculating minipage widths
\usepackage{etoolbox}
\makeatletter
\patchcmd\longtable{\par}{\if@noskipsec\mbox{}\fi\par}{}{}
\makeatother
\IfFileExists{footnotehyper.sty}{\usepackage{footnotehyper}}{\usepackage{footnote}}
\makesavenoteenv{longtable}
\usepackage{graphicx}
\makeatletter
\newsavebox\pandoc@box
\newcommand*\pandocbounded[1]{% scales image to fit in text height/width
  \sbox\pandoc@box{#1}%
  \Gscale@div\@tempa{\textheight}{\dimexpr\ht\pandoc@box+\dp\pandoc@box\relax}%
  \Gscale@div\@tempb{\linewidth}{\wd\pandoc@box}%
  \ifdim\@tempb\p@<\@tempa\p@\let\@tempa\@tempb\fi% select the smaller of both
  \ifdim\@tempa\p@<\p@\scalebox{\@tempa}{\usebox\pandoc@box}%
  \else\usebox{\pandoc@box}%
  \fi%
}
\def\fps@figure{htbp}
\makeatother

\NewDocumentCommand\citeproctext{}{}
\NewDocumentCommand\citeproc{mm}{%
  \begingroup\def\citeproctext{#2}\cite{#1}\endgroup}
\makeatletter
 \let\@cite@ofmt\@firstofone
 \def\@biblabel#1{}
 \def\@cite#1#2{{#1\if@tempswa , #2\fi}}
\makeatother
\newlength{\cslhangindent}
\newlength{\csllabelwidth}
\newenvironment{CSLReferences}[2] % #1 hanging-indent, #2 entry-spacing
 {\begin{list}{}{%
  \setlength{\itemindent}{0pt}
  \setlength{\leftmargin}{0pt}
  \setlength{\parsep}{0pt}
  \ifodd #1
   \setlength{\leftmargin}{\cslhangindent}
   \setlength{\itemindent}{-1\cslhangindent}
  \fi
  \setlength{\itemsep}{#2\baselineskip}}}
 {\end{list}}
\usepackage{calc}

\newcommand{\CSLLeftMargin}[1]{\parbox[t]{\csllabelwidth}{\strut#1\strut}}
\newcommand{\CSLRightInline}[1]{\parbox[t]{\linewidth - \csllabelwidth}{\strut#1\strut}}

\providecommand{\tightlist}{%
  \setlength{\itemsep}{0pt}\setlength{\parskip}{0pt}}

\usepackage{xcolor}
\definecolor{nhbblue}{HTML}{1F4E79}    % title / rules / links / footer
\definecolor{nhbabstract}{HTML}{EAF1FB} % shaded abstract-box background

\usepackage{tikz}
\usetikzlibrary{svg.path}
\definecolor{orcidlogocol}{HTML}{A6CE39}
\tikzset{
  orcidlogo/.pic={
    \fill[orcidlogocol] svg{M256,128c0,70.7-57.3,128-128,128C57.3,256,0,198.7,0,128C0,57.3,57.3,0,128,0C198.7,0,256,57.3,256,128z};
    \fill[white] svg{M86.3,186.2H70.9V79.1h15.4v48.4V186.2z}
                 svg{M108.9,79.1h41.6c39.6,0,57,28.3,57,53.6c0,27.5-21.5,53.6-56.8,53.6h-41.8V79.1z M124.3,172.4h24.5c34.9,0,42.9-26.5,42.9-39.7c0-21.5-13.7-39.7-43.7-39.7h-23.7V172.4z}
                 svg{M88.7,56.8c0,5.5-4.5,10.1-10.1,10.1c-5.6,0-10.1-4.6-10.1-10.1c0-5.6,4.5-10.1,10.1-10.1C84.2,46.7,88.7,51.3,88.7,56.8z};
  }
}
\makeatletter
\newcommand{\@OrigHeightRecip}{0.00390625}
\newlength{\@curXheight}
\DeclareRobustCommand\orcidlink[1]{%
\texorpdfstring{%
\setlength{\@curXheight}{\fontcharht\font`X}%
\href{https://orcid.org/#1}{\XeTeXLinkBox{\mbox{%
\begin{tikzpicture}[yscale=-\@OrigHeightRecip*\@curXheight,
xscale=\@OrigHeightRecip*\@curXheight,transform shape]
\pic{orcidlogo};
\end{tikzpicture}%
}}}}{}}
\makeatother

\usepackage{sectsty}
\sectionfont{\Large\bfseries}
\subsectionfont{\large\bfseries}
\subsubsectionfont{\normalsize\bfseries}

\usepackage[skins,breakable]{tcolorbox}
\newtcolorbox{nhbabstractbox}{%
  colback=nhbabstract, colframe=nhbabstract, boxrule=0pt,
  arc=2pt, left=10pt, right=10pt, top=8pt, bottom=8pt,
  boxsep=0pt, enhanced, breakable}

\usepackage[right]{lineno}
\newcommand*\patchAmsMathEnvironmentForLineno[1]{%
  \expandafter\let\csname old#1\expandafter\endcsname\csname #1\endcsname
  \expandafter\let\csname oldend#1\expandafter\endcsname\csname end#1\endcsname
  \renewenvironment{#1}%
    {\linenomath\csname old#1\endcsname}%
    {\csname oldend#1\endcsname\endlinenomath}}
\newcommand*\patchBothAmsMathEnvironmentsForLineno[1]{%
  \patchAmsMathEnvironmentForLineno{#1}%
  \patchAmsMathEnvironmentForLineno{#1*}}
\AtBeginDocument{%
  \patchBothAmsMathEnvironmentsForLineno{equation}%
  \patchBothAmsMathEnvironmentsForLineno{align}%
  \patchBothAmsMathEnvironmentsForLineno{flalign}%
  \patchBothAmsMathEnvironmentsForLineno{alignat}%
  \patchBothAmsMathEnvironmentsForLineno{gather}%
  \patchBothAmsMathEnvironmentsForLineno{multline}}

\usepackage{fancyhdr}
\usepackage{lastpage}
\providecommand{\nhbshorttitle}{}   % overridden in before-body.tex
\fancypagestyle{plain}{%
  \fancyhf{}%
  \fancyfoot[L]{\color{nhbblue}\itshape\small\nhbshorttitle}%
  \fancyfoot[R]{\color{nhbblue}\small\thepage/\pageref{LastPage}}}

\usepackage{etoolbox}
\usepackage{setspace}
\AtBeginEnvironment{longtable}{\small\setstretch{1.0}}

\makeatletter
\AtBeginDocument{%
  \renewcommand\paragraph{\@startsection{paragraph}{4}{\z@}%
    {-2.5ex \@plus -1ex \@minus -.2ex}%
    {1.2ex \@plus .2ex}%            positive afterskip => block heading
    {\normalfont\normalsize\bfseries}}%
  \renewcommand\subparagraph{\@startsection{subparagraph}{5}{\z@}%
    {-2.2ex \@plus -1ex \@minus -.2ex}%
    {0.8ex \@plus .2ex}%            positive afterskip => block heading
    {\normalfont\normalsize\itshape}}%
}
\makeatother

\usepackage{capt-of}
\makeatletter
\renewenvironment{figure}[1][]{%
  \par\addvspace{\intextsep}%
  \centering
  \def\@captype{figure}%
}{%
  \par\addvspace{\intextsep}%
}
\makeatother

\usepackage[export]{adjustbox}
\setkeys{Gin}{keepaspectratio,max width=\linewidth,max height=0.9\textheight}
\usepackage{newunicodechar}
\newunicodechar{α}{\ensuremath{\alpha}}
\newunicodechar{β}{\ensuremath{\beta}}
\newunicodechar{ε}{\ensuremath{\varepsilon}}
\newunicodechar{η}{\ensuremath{\eta}}
\newunicodechar{κ}{\ensuremath{\kappa}}
\newunicodechar{ρ}{\ensuremath{\rho}}
\newunicodechar{ω}{\ensuremath{\omega}}
\newunicodechar{Δ}{\ensuremath{\Delta}}
\newunicodechar{×}{\ensuremath{\times}}
\newunicodechar{−}{\ensuremath{-}}
\newunicodechar{±}{\ensuremath{\pm}}
\newunicodechar{≈}{\ensuremath{\approx}}
\newunicodechar{≤}{\ensuremath{\leq}}
\newunicodechar{≥}{\ensuremath{\geq}}
\newunicodechar{∈}{\ensuremath{\in}}
\newunicodechar{⁻}{\textsuperscript{\ensuremath{-}}}
\newunicodechar{⁰}{\textsuperscript{0}}
\newunicodechar{¹}{\textsuperscript{1}}
\newunicodechar{²}{\textsuperscript{2}}
\newunicodechar{³}{\textsuperscript{3}}
\newunicodechar{⁴}{\textsuperscript{4}}
\newunicodechar{⁵}{\textsuperscript{5}}
\newunicodechar{⁶}{\textsuperscript{6}}
\newunicodechar{⁷}{\textsuperscript{7}}
\newunicodechar{⁸}{\textsuperscript{8}}
\newunicodechar{⁹}{\textsuperscript{9}}
\newunicodechar{—}{\textemdash{}}
\newunicodechar{£}{\pounds{}}
\makeatletter
\@ifpackageloaded{caption}{}{\usepackage{caption}}
\AtBeginDocument{%
\ifdefined\contentsname
  \renewcommand*\contentsname{Table of contents}
\else
  \newcommand\contentsname{Table of contents}
\fi
\ifdefined\listfigurename
  \renewcommand*\listfigurename{List of Figures}
\else
  \newcommand\listfigurename{List of Figures}
\fi
\ifdefined\listtablename
  \renewcommand*\listtablename{List of Tables}
\else
  \newcommand\listtablename{List of Tables}
\fi
\ifdefined\figurename
  \renewcommand*\figurename{Figure}
\else
  \newcommand\figurename{Figure}
\fi
\ifdefined\tablename
  \renewcommand*\tablename{Table}
\else
  \newcommand\tablename{Table}
\fi
}
\@ifpackageloaded{float}{}{\usepackage{float}}
\floatstyle{ruled}
\@ifundefined{c@chapter}{\newfloat{codelisting}{h}{lop}}{\newfloat{codelisting}{h}{lop}[chapter]}
\floatname{codelisting}{Listing}

\makeatother
\makeatletter
\@ifpackageloaded{caption}{}{\usepackage{caption}}
\@ifpackageloaded{subcaption}{}{\usepackage{subcaption}}
\makeatother
\usepackage{bookmark}
\IfFileExists{xurl.sty}{\usepackage{xurl}}{} % add URL line breaks if available
\hypersetup{
  pdftitle={Assessing mentalization in humans and large language models},
  pdfauthor={Aamir Sohail; Xintong Zhong; Arkady Konovalov; Patricia L. Lockwood; Lei Zhang},
  pdfkeywords={mentalization, large language models, artificial
intelligence, computational modeling},
  colorlinks=true,
  linkcolor={nhbblue},
  filecolor={Maroon},
  citecolor={nhbblue},
  urlcolor={nhbblue},
  pdfcreator={LaTeX via pandoc}}

\title{Assessing mentalization in humans and large language models}
\author{Aamir Sohail \and Xintong Zhong \and Arkady
Konovalov \and Patricia L. Lockwood \and Lei Zhang}
\date{26 August 2026}
\begin{document}

% ---- short title for the running footer --------------------
\renewcommand{\nhbshorttitle}{Assessing mentalization in humans and
large language models}

% ============================================================
%  Title block  (Nature Human Behaviour preprint style)
% ============================================================
\thispagestyle{fancy}
\begingroup
\raggedright

{\color{nhbblue}\fontsize{22}{26}\selectfont\bfseries Assessing
mentalization in humans and large language models\par}
\vspace{1.0em}

% ---- Authors: name, corresponding *, superscript affil nums -
{\fontsize{14}{19}\selectfont\bfseries
Aamir
Sohail\hspace{0.15em}\orcidlink{0009-0000-6584-4579}\hspace{0.15em}\textsuperscript{1,2,3*}, Xintong
Zhong\hspace{0.15em}\hspace{0.15em}\textsuperscript{4}, Arkady
Konovalov\hspace{0.15em}\orcidlink{0000-0002-9448-6659}\hspace{0.15em}\textsuperscript{1,2}, \\Patricia
L.
Lockwood\hspace{0.15em}\orcidlink{0000-0001-7195-9559}\hspace{0.15em}\textsuperscript{1,2,3,5,6}, Lei
Zhang\hspace{0.15em}\orcidlink{0000-0002-9586-595X}\hspace{0.15em}\textsuperscript{1,2,3,5*}\par}
\vspace{0.9em}

% ---- Numbered affiliation list (italic) --------------------
{\small\itshape
\noindent \textsuperscript{1}Centre for Human Brain Health, University
of Birmingham, Birmingham, UK.\par
\noindent \textsuperscript{2}School of Psychology, University of
Birmingham, Birmingham, UK.\par
\noindent \textsuperscript{3}Institute for Mental Health, University of
Birmingham, Birmingham, UK.\par
\noindent \textsuperscript{4}Department of Psychology, Virginia
Tech, Blacksburg, USA.\par
\noindent \textsuperscript{5}Centre for Developmental
Science, University of Birmingham, Birmingham, UK.\par
\noindent \textsuperscript{6}Department of Experimental
Psychology, University of Oxford, Oxford, UK.\par
}
\vspace{0.8em}

% ---- Corresponding author emails (semicolon-separated) -----
\newif\ifnhbfirstcorr \nhbfirstcorrtrue
{\small\noindent\textbf{*\,Corresponding author emails:}~\ifnhbfirstcorr\nhbfirstcorrfalse\else;\space\fi{}axs2210@student.bham.ac.uk\ifnhbfirstcorr\nhbfirstcorrfalse\else;\space\fi{}l.zhang.13@bham.ac.uk\par}

\endgroup
\vspace{1.6em}

% ============================================================
%  Abstract  (shaded box)  +  keywords
% ============================================================
\noindent{\Large\bfseries Abstract\par}
\vspace{0.5em}
\begin{nhbabstractbox}
Mentalization - the ability to infer others' beliefs and intentions to
guide one's own choices - is a key cognitive function underlying human
social interactions. Large language models (LLMs) demonstrate behaviour
consistent with humans on theory-of-mind tasks, yet whether these models
can guide adaptive behaviour through mentalization is unknown. Here we
use two economic games with cognitive computational modeling to uncover
the latent strategies underlying mentalization in LLMs. We tested
individual LLM agents across four model families, DeepSeek, GPT-4.1,
GPT-5 and Gemini 2.0 Flash (N = 2,099), against opponents of varying
sophistication and examined whether a prompting strategy designed to
elicit strategic reasoning improved performance. We benchmarked results
against human participants (N = 251) as a comparative measure. Across
both games, LLMs showed clear behavioural and computational signatures
of mentalizing that differed markedly by model provider and size.
Strategic prompting generally improved performance by inducing more
sophisticated reasoning, yet the extent of the benefit differed across
the two tasks. Last, GPT-5 agents flexibly adapted their recursive depth
of reasoning to increasingly sophisticated opponents, demonstrating
superior performance to human participants. Collectively, we demonstrate
different capacities for mentalization across LLMs, and highlight
cognitive computational modeling as a formal method for assessing
comparative intelligence across humans and machines.
\end{nhbabstractbox}

\vspace{0.6em}
{\noindent\textbf{Keywords:}~\itshape mentalization, large language
models, artificial intelligence, computational modeling\par}

\vspace{1.2em}

% Begin continuous right-margin line numbering for the body.
% arXiv forbids margin line numbers, so this is gated on the `linenumbers`
% metadata key: true for journal formats, false for the nhb-arxiv variant.

\setstretch{1.15}
\section{Main}\label{sec-main}

Mentalization is a cognitive process associated with interpreting the
behaviour of others as the result of latent mental states, beliefs and
emotions
{[}\citeproc{ref-fonagy2018}{1}--\citeproc{ref-freeman2016}{3}{]}. In
humans, mentalization shapes decision-making in social contexts by
predicting the actions of others and adjusting one's own behaviour
accordingly
{[}\citeproc{ref-apperly2012}{4}--\citeproc{ref-tamir2018}{10}{]}.
Evidence also has suggested the presence of mentalization in a select
range of non-human animal species, including dogs, corvids, chimpanzees
and gorillas
{[}\citeproc{ref-berke2025}{11}--\citeproc{ref-taylor2014}{20}{]}
reflecting an evolved capacity for social intelligence.

Beyond humans and other animals, a topic of significant recent
development concerns whether generative artificial intelligence (AI)
systems, such as large language models (LLMs) demonstrate behaviour
consistent with mentalizing. Researchers have turned to experimental
methods applied within the fields of psychology, cognitive science and
neuroscience to uncover features of behaviour and reasoning abilities in
LLMs {[}\citeproc{ref-brady2025}{21}--\citeproc{ref-zador2023}{30}{]}.
By converting behavioural tasks to text which are submitted as prompts,
assessments of learning and decision-making - typically used among human
participants - can be directly applied to LLMs, providing deeper
computational insights beyond superficial measures of choice
{[}\citeproc{ref-binz2023}{31}--\citeproc{ref-sucholutsky2025}{37}{]}.

Despite being a type of neural network trained upon large quantities of
text-based data {[}\citeproc{ref-vaswani2017}{38}{]}, LLMs demonstrate
emergent properties synonymous with reasoning and decision-making in
humans {[}\citeproc{ref-brady2025}{21}, \citeproc{ref-binz2023}{31},
\citeproc{ref-abdulhai2023}{39}--\citeproc{ref-yax2024}{48}{]}. In
humans, mentalization is conventionally assessed using role-based tasks,
stories necessitating the inference of fictional characters' beliefs
{[}\citeproc{ref-wellman2001}{49}, \citeproc{ref-wimmer1983}{50}{]}.
Large language models are similarly able to correctly interpret mental
states and suggest appropriate actions, often performing at or above
human level in such tasks
{[}\citeproc{ref-bubeck2023}{51}--\citeproc{ref-zhou2023}{57}{]}.
However, LLMs fail at simple modifications to these assessments
{[}\citeproc{ref-ullman2023}{58}{]} and struggle in theory-of-mind (ToM)
tasks necessitating complex reasoning and planning
{[}\citeproc{ref-saritas2025}{55}, \citeproc{ref-attanasio2024}{59},
\citeproc{ref-moore2025}{60}{]}, casting doubt as to whether successes
reflect genuine mentalization or shallow heuristics
{[}\citeproc{ref-marchetti2025}{61}--\citeproc{ref-shapira2023}{63}{]}.
As a result, it has been recently argued that the role-based tasks
commonly used may not be suitable for assessing complex cognitive
processes {[}\citeproc{ref-marchetti2025}{61}{]}, by only describing
superficial elements of behaviour
{[}\citeproc{ref-holtzman2025}{64}--\citeproc{ref-wagner2025}{69}{]}
which may be solved without requiring the explicit, human-like
simulation of mental states {[}\citeproc{ref-lu2025}{70}{]}. Therefore,
whilst the observed choices between human and LLMs may be identical,
this does not necessarily reflect the same inferential or cognitive
process {[}\citeproc{ref-hu2025}{65}, \citeproc{ref-ku2025}{66},
\citeproc{ref-blank2023}{71}, \citeproc{ref-connell2024}{72}{]}.

Contrasting with story-based assessments, economic games present a
formal methodology for assessing strategic social behaviour by examining
distinct behavioural and computational mechanisms
{[}\citeproc{ref-camerer2003}{73}--\citeproc{ref-zhang2020a}{79}{]}. In
these games, participants are directly part of the interaction rather
than a third-party observer, and are often tasked with outplaying an
opponent {[}e.g., the Inspection Game;
{[}\citeproc{ref-hampton2008}{80}{]}{]}. Importantly, this context
allows researchers to test recursive mentalization at different levels
by experimentally manipulating the reasoning depth of the opponent.
Under this framework, a player may attempt to estimate the opponent's
historical choice distribution and form first-order beliefs, or -- if
anticipating that their actions are being tracked - form second-order
beliefs and modify their strategy according to her opponent's reaction
{[}\citeproc{ref-konovalov2026}{77}, \citeproc{ref-camerer2004}{81},
\citeproc{ref-devaine2014}{82}{]}. Accordingly, this reasoning process
can also reach higher levels. Conventionally used to assess behaviour
among human participants, economic games have recently been employed to
objectively measure specific biases and patterns of decision-making in
LLMs {[}\citeproc{ref-brady2025}{21}, \citeproc{ref-binz2023}{31},
\citeproc{ref-abdulhai2023}{39}--\citeproc{ref-sun2025a}{46},
\citeproc{ref-yax2024}{48},
\citeproc{ref-brookins2024}{83}--\citeproc{ref-zhang2024}{92}{]}, which
demonstrate emergent properties of deliberative thought. Furthermore,
the performance of LLMs on these tasks is often improved using targeted
prompting strategies -- such as chain-of-thought (CoT) --
{[}\citeproc{ref-chen2025a}{93}--\citeproc{ref-wei2022}{95}{]}, designed
to elicit step-by-step reasoning indicative of human decision-making
{[}\citeproc{ref-akata2025}{40}, \citeproc{ref-yax2024}{48},
\citeproc{ref-anglin2025}{96}--\citeproc{ref-schoenegger2025}{100}{]}.

In humans, choice behaviour can be formally analysed using cognitive
computational modeling, a theory-driven quantification ascribing latent
cognitive processes and computational parameters to observed actions
{[}\citeproc{ref-farrell2018}{101}--\citeproc{ref-wilson2019}{105}{]}.
In the context of dynamic social behaviour, these parameters can
additionally be assessed trial-by-trial to uncover temporal patterns of
cognition. Concerning mentalization, computational models have
highlighted the capacity for recursive mentalization in humans
{[}\citeproc{ref-hampton2008}{80}, \citeproc{ref-hill2017}{106}{]} who
are able to dynamically adapt their recursive depth to the inferred
strategy of the opponent, i.e., \emph{adaptive} mentalization
{[}\citeproc{ref-buergi2026}{107}{]}. Cognitive models also offer a
mechanistic formalization of behaviour across species
{[}\citeproc{ref-redish2021}{108}, \citeproc{ref-robbins2019}{109}{]},
solely requiring choice data to infer upon cognitive processes.
Subsequently, computational modeling has recently been endorsed in AI
and LLM research as a formal approach to infer the internal processes of
AI systems from observed choices {[}\citeproc{ref-palminteri2025}{110},
\citeproc{ref-taschereau-dumouchel2026}{111}{]}. For example, this
approach has uncovered latent parameters underlying learning biases in
LLMs, by fitting reinforcement learning (RL) models
{[}\citeproc{ref-coda-forno2024}{112}--\citeproc{ref-schubert2024}{114}{]}.
To date, a comparative assessment of mentalization across humans and
LLMs using strategic games has yet to be undertaken, despite a growing
interest towards developing `human-like' machine intelligence
{[}\citeproc{ref-lake2017}{26}, \citeproc{ref-collins2024}{115}{]}.

In the current study, we employ two commonly used economic games -- the
inspection game and rock-paper-scissors - to uncover the behavioural
signatures and latent parameters of recursive mentalization in large
language models. We include several popular LLMs that were subjected to
the experimental tasks as participants (total N = 2,099), and examine
the impact of a simple prompting strategy -- Social Chain-of-Thought
(SCoT) -- previously shown to improve LLM performance in economic games
{[}\citeproc{ref-akata2025}{40}{]}. We also compare their performance to
human participants (total N = 251; mean age = 24.48 ± 3.81 years; range
18--36) across both tasks, analysing differences in recursive
mentalization across humans and LLMs. In doing so, we move beyond
conventional descriptive assessments and provide a more formal
understanding of machine intelligence.

\section{Results}\label{sec-results}

For our study, we selected two repeated-interaction games commonly used
to assess strategic play in humans, the Inspection game and
rock-paper-scissors (RPS). To assess recursive mentalization, we
employed the inspection game {[}\citeproc{ref-hampton2008}{80},
\citeproc{ref-hill2017}{106}{]}, a repeated non-cooperative game where
two players assume the roles of an ``employee'' and an ``employer'' and
make one of two choices simultaneously before receiving a payoff (Fig.
1a). On each trial, the employee chooses whether to ``work'' or
``shirk,'' and the employer chooses whether to ``inspect'' or ``not
inspect'' the employee. In our version, the participants (i.e.,
human/LLM) always played the role of the employee. Subsequently, the
payoff matrix is specified such that participants earn points by
adapting to the opponent's predicted choice. Participants played against
a computer algorithm (designated as the employer, operating at level
\emph{k} = 2, second-order belief) designed to make adaptive choices to
participants' expectations over the course of the task. Therefore, the
total payoff received reflects the participants' capacity for recursive
reasoning. We also tested participants' ability for \emph{adaptive}
mentalization; the ability to flexibly adapt the level of recursive
mentalization to opponents of different sophistication. For this, we
employed a zero-sum RPS game previously used with human participants
{[}\citeproc{ref-buergi2026}{107}{]}. In contrast to the inspection
game, which is played against a fixed algorithm across all trials,
participants in the RPS game played against an algorithm programmed to
make choices in accordance with a specific reasoning level \emph{k},
where \emph{k} ∈ \{0, 1, 2\}. On each trial, participants received a
single point for winning, lost a single point for losing and received no
points in the case of a tie. Participants played a block of 40 trials
against each \emph{k}-level (a total of 120 trials across 3 blocks) with
blocks randomised and counterbalanced across participants (Fig. 1b).

Data were collected in both human participants (inspection game: n = 67;
RPS: n = 184) and LLMs (inspection game: all groups n ≥ 49; RPS: all
groups n ≥ 188; see Methods). Human participants in both tasks were
provided with instructions that they would be playing against a real
other person, but were in fact playing against the respective computer
algorithm. Human data for the inspection game were collected in person,
and online for the RPS game (see Methods). Tasks were appropriately
re-coded into a text-based format for API inference in LLMs (see
Methods/SI) (Fig. 1c). Converted tasks retained identical payoff rules,
opponent configurations and trial lengths, however, as LLMs demonstrate
bias in the conventional RPS game {[}\citeproc{ref-vidler2025}{116}{]},
choices were re-coded in the RPS game to letters J/Q/Z that did not
carry semantic meaning for LLMs; an established protocol commonly used
to control for text biases {[}\citeproc{ref-akata2025}{40}{]}. To
uncover the possible effects of strategic prompting on mentalization,
LLMs were also prompted with Social Chain-of-Thought (SCoT)
{[}\citeproc{ref-akata2025}{40}{]}, a prompting scheme where LLMs are
specifically asked to make a prediction on the opponents action in each
trial before making their choice. The following LLMs were selected for
participation in the respective tasks. Inspection game: Gemini,
DeepSeek, GPT-4.1, GPT-5; RPS: DeepSeek, GPT-5
(Table~\ref{tbl-overview}). The selections were made considering
practical constraints with monetary costs for running the study and the
availability of specific models at the time of data collection (4/8/2025
-- 14/10/2025). Where possible, LLMs were tested under both conditions
(conventional prompt + SCoT), with the deliberate exception of GPT-5
which natively supports deliberative reasoning
{[}\citeproc{ref-singh2026}{117}{]}.

\begin{figure}

\centering{

\includegraphics[width=0.78\linewidth,height=\textheight,keepaspectratio]{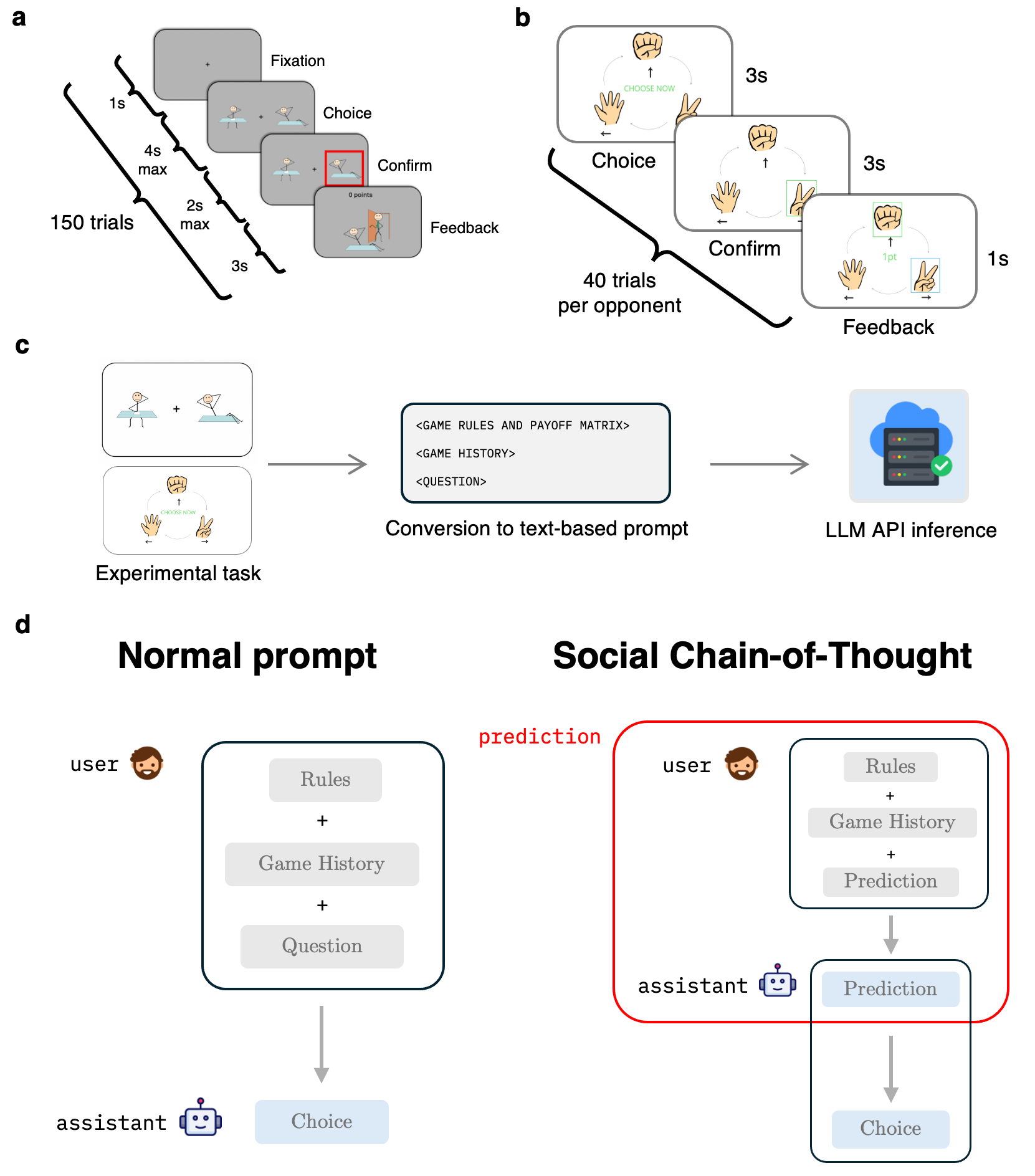}

}

\caption{\label{fig-design}\textbf{Experimental design for humans and
large language models.} \textbf{(a)} For the inspection game, on each
trial human participants were firstly presented with a fixation cross (1
second) and had a maximum of 4 seconds to make their choice, followed by
the choice confirmation (max 2 seconds). They were then provided
feedback based on the choice of the opponent (3 seconds). Across a
single block of 150 trials, humans played against a sophisticated
algorithm designed to engage in the `Influence play' signature of
reasoning. \textbf{(b)} In the RPS task, humans on each trial had a
maximum of 3 seconds to make their choice, after which they received
feedback for 1 second. Participants played 3 blocks of 40 trials against
an adaptive algorithm which made decisions reflecting different levels
of mentalization, from zero-order (k=0) to second-order (k=2). The block
order was counterbalanced and randomised across participants.
\textbf{(c)} For large language model (LLM) participants, each task was
converted into a text-based prompt, matching the original task where
possible. These were then submitted to the LLMs for API inference.
\textbf{(d)} Prompting conditions across both tasks. In the normal
prompt, the task rules and choice history were presented along with a
statement asking for the agent's choice. In the SCoT condition, the
agent was instead asked to predict the opponent's choice. This
prediction was then used to select the corresponding rational choice.
Icons from Flaticons.com.}

\end{figure}%

\begin{longtable}[]{@{}
  >{\raggedright\arraybackslash}p{(\linewidth - 6\tabcolsep) * \real{0.2500}}
  >{\raggedright\arraybackslash}p{(\linewidth - 6\tabcolsep) * \real{0.2500}}
  >{\raggedright\arraybackslash}p{(\linewidth - 6\tabcolsep) * \real{0.2500}}
  >{\raggedright\arraybackslash}p{(\linewidth - 6\tabcolsep) * \real{0.2500}}@{}}
\caption{\textbf{An overview of the algorithms and large language models
used in the study.} \emph{Note.} CHASE = Cognitive Hierarchy Assessment;
EWA = Experience Weighted Attraction; ME = Mixed-equilibrium; RL =
Reinforcement Learning; ToMk = Theory-of-Mind
\emph{k}.}\label{tbl-overview}\tabularnewline
\toprule\noalign{}
\begin{minipage}[b]{\linewidth}\raggedright
\textbf{Inspection game}
\end{minipage} & \begin{minipage}[b]{\linewidth}\raggedright
\end{minipage} & \begin{minipage}[b]{\linewidth}\raggedright
\textbf{Rock paper scissors}
\end{minipage} & \begin{minipage}[b]{\linewidth}\raggedright
\end{minipage} \\
\midrule\noalign{}
\endfirsthead
\toprule\noalign{}
\begin{minipage}[b]{\linewidth}\raggedright
\textbf{Inspection game}
\end{minipage} & \begin{minipage}[b]{\linewidth}\raggedright
\end{minipage} & \begin{minipage}[b]{\linewidth}\raggedright
\textbf{Rock paper scissors}
\end{minipage} & \begin{minipage}[b]{\linewidth}\raggedright
\end{minipage} \\
\midrule\noalign{}
\endhead
\bottomrule\noalign{}
\endlastfoot
\textbf{Participants} & Human participants & \textbf{Participants} &
Human participants \\
& DeepSeek / DeepSeek-SCoT & & DeepSeek / DeepSeek-SCoT \\
& Gemini / Gemini-SCoT & & GPT-5 \\
& GPT-4.1 / GPT-4.1-SCoT & & \\
& GPT-5 & & \\
\textbf{Models} & Influence play & \textbf{Models} & CHASE \\
& Fictitious play & & Reward learner \\
& RL & & Self-tuning EWA \\
& Mixed-equilibrium (ME) & & Full EWA \\
& & & Fictitious play \\
& & & ToMk \\
\end{longtable}

\subsection{Assessing mentalization using the inspection
game}\label{sec-results-inspection}

The LLMs altogether consisted of seven groups: gemini-2.0-flash
(Gemini), gemini-2.0-flash with SCoT (Gemini-SCoT), deepseek-chat
(DeepSeek), deepseek-chat with SCoT (DeepSeek-SCoT), gpt-4.1-2025-04-14
(GPT-4.1), gpt-4.1-2025-04-14 with SCoT (GPT-4.1-SCoT), and
gpt-5-2025-08-07 (GPT-5). We first tested how performance on the
inspection game differed across each LLM (all n \textgreater{} 48) and
prompting strategy compared with the human group (n = 67). As LLM groups
met normality assumptions but presented unequal variance, Welch's
independent samples t-tests were used, with Bonferroni correction
applied across the seven comparisons representing the LLM model type ×
prompting strategy combination (α = 0.007). Altogether, most
combinations earned significantly different payoffs from humans.
Specifically, Gemini (M\textsubscript{payoff} = 14.2;
t\textsubscript{(112.01)} = -36.6, p = 3.58×10\textsuperscript{-64}, g =
-6.40, 95\% CI {[}-7.29, -5.49{]}), Gemini-SCoT (M\textsubscript{payoff}
= 22.4; t\textsubscript{(109.42)} = -13.7, p =
1.59×10\textsuperscript{-25}, g = -2.37, 95\% CI {[}-2.84, -1.89{]}),
DeepSeek (M\textsubscript{payoff} = 18.0; t\textsubscript{(110.59)} =
-26.5, p = 1.16×10\textsuperscript{-49}, g = -4.59, 95\% CI {[}-5.28,
-3.89{]}), DeepSeek-SCoT (M\textsubscript{payoff} = 25.1;
t\textsubscript{(106.59)} = -6.04, p = 2.32×10\textsuperscript{-8}, g =
-1.04, 95\% CI {[}-1.42, -0.65{]}) and GPT-4.1 (M\textsubscript{payoff}
= 18.5; t\textsubscript{(95.42)} = -27.2, p =
8.7×10\textsuperscript{-47}, g = -4.57, 95\% CI {[}-5.26, -3.88{]}), all
earned significantly less than humans. The only exceptions were
GPT-4.1-SCoT (M\textsubscript{payoff} = 27.8; t\textsubscript{(114.78)}
= 1.74, p = 0.085, g = 0.31, 95\% CI {[}-0.06, 0.68{]}) and GPT-5
(M\textsubscript{payoff} = 26.6; t\textsubscript{(112.18)} = -1.55, p =
0.123, g = -0.26, 95\% CI {[}-0.57, 0.05{]}), which did not
significantly differ from humans. We next sought to understand how the
choice of LLM and the prompting strategy jointly affected performance.
For the three models tested under both strategies (DeepSeek, Gemini,
GPT-4.1), we fitted a 2 (prompt) × 3 (model) factorial ANOVA to mean
payoff per trial. As cell variances were unequal (Levene's
F\textsubscript{(5,~293)} = 3.79, p = 0.002), we report Type-III Wald
F-tests. We found significant main effects of model
(F\textsubscript{(2,~293)} = 286.95, p = 9.62×10\textsuperscript{-70},
partial η² = 0.68, 95\% CI {[}0.62, 0.72{]}), and prompt
(F\textsubscript{(1,~293)} = 2501.89, p = 1.57×10\textsuperscript{-145},
partial η² = 0.90, 95\% CI {[}0.88, 0.91{]}), with a significant model ×
prompt interaction (F\textsubscript{(2,~293)} = 14.96, p =
6.53×10\textsuperscript{-7}, partial η² = 0.09, 95\% CI {[}0.04,
0.16{]}). Whilst every model earned significantly higher payoffs under
SCoT (Fig. 2a), the magnitude of this benefit varied, being smallest for
DeepSeek (Hedges' g = −5.29, 95\% CI {[}−6.12, −4.45{]}, p =
1.95×10\textsuperscript{-46}), intermediate for Gemini (g = −5.77,
{[}−6.67, −4.87{]}, p = 3.64×10\textsuperscript{-49}) and largest for
GPT-4.1 (g = −6.31, {[}−7.28, −5.35{]}, p =
5.26×10\textsuperscript{-46}). GPT-5, tested under normal prompting
alone, outperformed all other normally-prompted models
(F\textsubscript{(3,~121.62)} = 821, p = 1.61×10\textsuperscript{-80},
ω² = 0.92, 95\% CI {[}0.90, 0.93{]}; Games--Howell all p \textless{}
0.001 except DeepSeek vs.~GPT-4.1, p = 0.17). Therefore, whilst both
model and SCoT prompting impacted task performance, where SCoT benefited
every model, the size of that benefit was model-dependent.

We then examined whether the predictions for the opponent's choice made
by each model under SCoT prompting differed from chance (50\%) (Fig.
2b). One-sample t-tests revealed significant differences in predictive
accuracy, with GPT-4.1-SCoT predicting the opponent's choice
significantly above chance level (t\textsubscript{(49)} = 10.98, p =
8.33×10\textsuperscript{-15}, d = 1.55), Gemini-SCoT predicting
significantly below chance (t\textsubscript{(49)} = -13.32, p =
6.56×10\textsuperscript{-18}, d = -1.88), while DeepSeek-SCoT did not
differ significantly from chance (t\textsubscript{(49)} = 0.63, p =
0.53, d = 0.09) (Fig. 2b). Furthermore, examining switch frequency -
trials where the subject changed choices from the previous trial -
revealed differences across groups (Fig. 2c). Across groups, the mean
switch frequency also significantly correlated with the mean payoff (r =
0.735, 95\% CI {[}0.063, 0.948{]}, t\textsubscript{(6)} = 2.657, p =
0.038, R² = 0.540) showing that strategic adaptation was crucial for
performance against the opponent (Fig. 2d). Social-Chain-of-Thought
therefore presents a simple, robust prompting method for improving
performance among LLMs in the inspection game by enhancing predictive
reasoning towards the actions of the opponent. However, this improvement
was scaled to the model's performance under normal prompting.

\newpage

\begin{figure}

\centering{

\includegraphics[width=1\linewidth,height=\textheight,keepaspectratio]{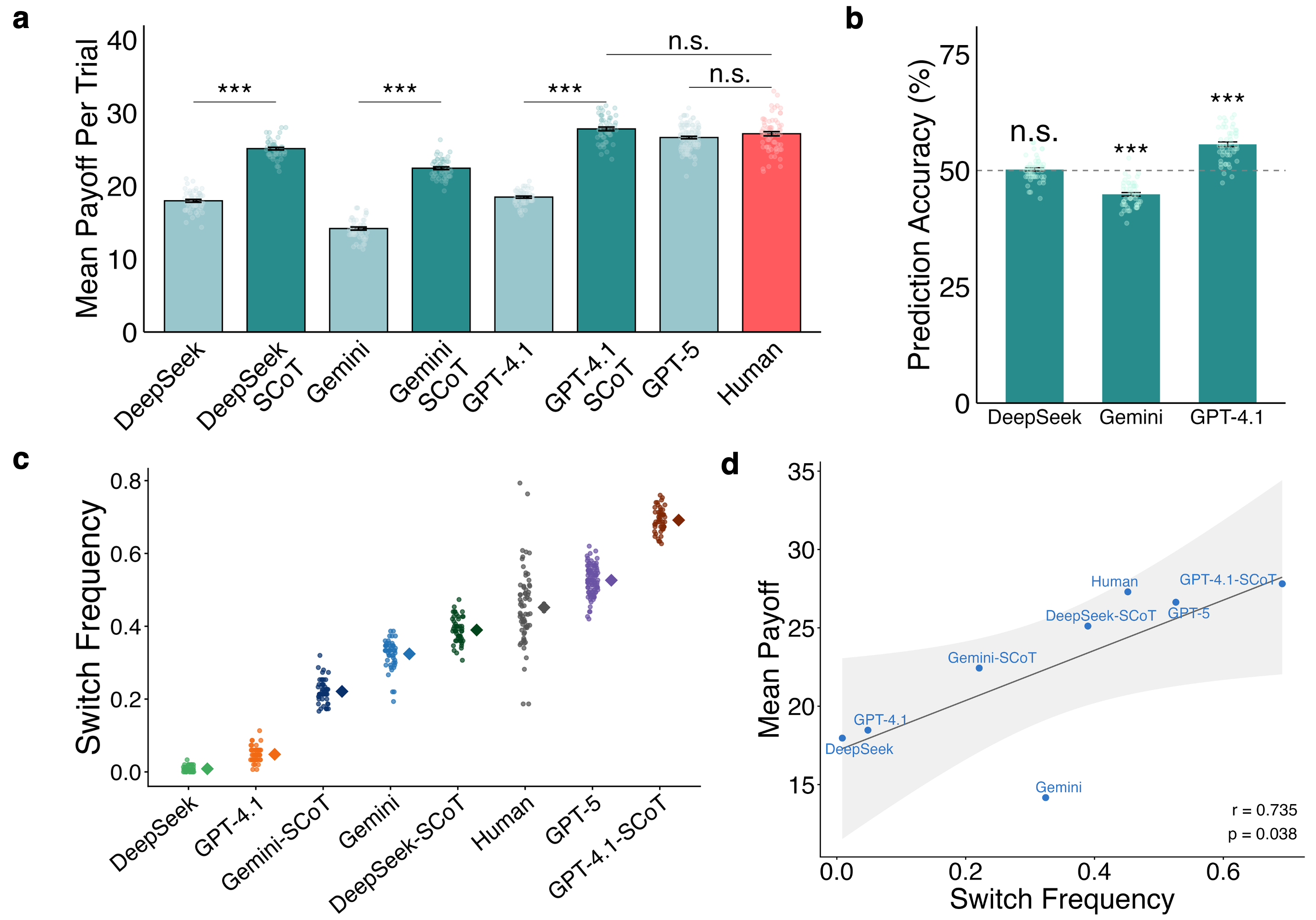}

}

\caption{\label{fig-inspection-behav}\textbf{Behavioural results for
humans and LLMs in the inspection game.} \textbf{(a)} Social
Chain-of-Thought (SCoT) robustly improved performance on the task across
LLMs, as measured by the mean payoff achieved per trial. Human
participants did not significantly differ in their performance on the
inspection game with GPT-5 and GPT-4.1-SCoT agents. \textbf{(b)} Among
SCoT-prompted LLMs, GPT-4.1-SCoT predicted the opponent's choice
significantly above chance, Gemini-SCoT significantly below chance,
whilst DeepSeek-SCoT did not deviate significantly from chance level in
its predictions. \textbf{(c)} Switch frequencies varied across human and
LLM groups, with the greatest between-subject spread observed among
humans. \textbf{(d)} Mean payoff and switch frequency are positively
correlated across LLMs and humans. Asterisks represent significant
differences (* p\textless0.05, ** p\textless0.01, *** p\textless0.001).}

\end{figure}%

Having established a behavioural difference between SCoT and
normally-prompted LLM groups, we next tested whether these behavioural
changes manifested computationally through altered reasoning mechanisms
following the established computational framework of this task
{[}\citeproc{ref-hampton2008}{80}{]}. To do so, we fit a computational
model (`influence play') that formalizes both how players react to their
opponent's past choices (first-order beliefs) and how they anticipate
the influence of their own choice on the opponent's behaviour (i.e.,
second-order beliefs). We also fit two simpler complementary models: a
reinforcement learning (RL; zero-order beliefs) model that learns action
values from payoffs without representing the opponent, and a model
representing first-order mentalization (`fictitious play'; first-order
beliefs). We assessed the models both through their predictive accuracy,
by assessing which model best predicts observed choices, and their
generative accuracy, by determining whether the winning model reproduces
the observed behaviour.

To determine predictive accuracy, we fitted each model to the
behavioural data of participants within each group under the
hierarchical Bayesian framework {[}\citeproc{ref-zhang2020a}{79},
\citeproc{ref-sohail2025a}{118}{]}, in which each participant's
parameters were drawn from a group-level distribution. We compared the
three cognitive models alongside a mixed-equilibrium (ME) strategy
{[}\citeproc{ref-hill2017}{106}{]} under which choices feature no
trial-by-trial learning, using the leave-one-out information criterion
(LOOIC) where a lower score indicates better fit (see Methods).

Model comparison (Fig. 3a) indicated that human participants employed
second-order mentalization, with the influence play model providing the
best fit to the data (LOOIC = 12236; ΔELPD vs.~mixed-equilibrium
reference = +60.0, SE = 25.6). Among LLMs, SCoT prompting consistently
shifted the winning model toward more sophisticated opponent-tracking.
Specifically, DeepSeek shifted from reinforcement learning (LOOIC = 708;
ΔELPD = +99.5, SE = 25.1) to fictitious play (LOOIC = 5387; ΔELPD =
+2306.8, SE = 47.7), while GPT-4.1 shifted from reinforcement learning
(LOOIC = 2557; ΔELPD = +147.7, SE = 29.0) to influence play (LOOIC =
8829; ΔELPD = +721.1, SE = 34.1). For Gemini, no learning model
outperformed the mixed-equilibrium reference under normal prompting,
whereas SCoT prompting produced a clear influence play winner (LOOIC =
5589; ΔELPD = +1269.5, SE = 51.2). Finally, GPT-5 was best fit by the
influence play model (LOOIC = 16782; ΔELPD = +1152.6, SE = 39.8),
demonstrating an ability for sophisticated reasoning without being
explicitly cued to do so (see Appendix for full model fits).

We next tested the direct mapping of the influence-play model's
parameters with behaviour by regressing model-free behavioural measures
on the fitted first-order learning rate (η) and second-order weight (κ)
within each group (see Table S3 for the complete per-group
coefficients). Following previous results among human participants
{[}\citeproc{ref-hill2017}{106}{]}, we anticipated three relationships
if the parameters map onto behaviour as intended. Namely, a higher
second-order weight (κ) should produce more switching, a higher
first-order learning rate (η) should produce stronger tracking of the
employer's most recent action, and a higher κ should produce a higher
mean payoff. This pattern was indeed realised in the human group (Fig.
3b), where all three predicted relationships held and were statistically
significant. The parameter representing second-order beliefs (κ) was
strongly associated with switching (b = 0.101, t\textsubscript{(64)} =
7.76, 95\% CI {[}0.075, 0.128{]}, p = 8.5×10\textsuperscript{-11}) and
task performance (b = 0.027, t\textsubscript{(64)} = 3.86, 95\% CI
{[}0.013, 0.041{]}, p = 2.7×10\textsuperscript{-4}), whilst the
first-order learning rate (η) was associated with adaptation to the
employer's last move (b = 0.163, t\textsubscript{(64)} = 6.85, 95\% CI
{[}0.115, 0.210{]}, p = 3.4×10\textsuperscript{-9}). However, across the
LLM groups, this mapping was weaker and only partial (see Table S3). For
LLM groups, comprising agents running the same model under a fixed
prompt, the between-subject variance was minimal, reflecting homogeneity
among parameter estimates. Therefore, whilst the switching and
opponent-tracking relationships often reached significance in individual
LLM groups, this was a likely consequence of less variation in the
parameter estimates. Conversely, the human group represented a
heterogeneous sample which carried genuine between-subject variance
compared to LLM groups. Ultimately these results remain interpretable
only for human participants, and reflect challenges concerning the
homogeneity of parameter estimates among LLM agents.

Finally, we assessed the generative fit of each winning model by
simulating choices from its own generated history and comparing the
behaviour to several choice patterns from the observed data: the
probability of working, switch rate and mean payoff
(Table~\ref{tbl-ig-ppc}; see Methods for full details). To do so, for
each model, we took the posterior-predictive distribution of the
discrepancy between the simulated and observed statistic, testing for
the percentage of the resulting distribution which fell within ±0.05 of
the shared {[}0,1{]} range using a Region of Practical Equivalence
(ROPE). Following recommended guidelines
{[}\citeproc{ref-kruschke2018}{119},
\citeproc{ref-makowski2019}{120}{]}, we accepted a ROPE statistic when
more than 97.5\% of the distribution fell within the range.

Building on earlier results, the influence play model also demonstrated
strong generative accuracy within human subjects, reproducing all three
statistics within the ROPE (Fig. 3c). On-the-other-hand, the LLM groups
differed along prompting condition. Where SCoT prompting was applied,
the winning model - influence play for Gemini and GPT-4.1, and
fictitious play for DeepSeek - mostly produced simulated behaviour
within the ROPE. Under normal prompting, by contrast, the groups whose
selected model was reinforcement learning (DeepSeek and GPT-4.1)
generally failed to accurately reproduce their observed behaviour.
Plotting the cumulative probability of work, p(Work), across trials
(Fig. 3d) revealed that the RL model was found to slowly drift the
agent's probability of working upwards, while the observed rate among
these two groups remained near zero, signalling its poor generative fit
and underlying choice homogeneity. Finally, GPT-5, an influence player
without any Chain-of-Thought cue, demonstrated a mixed accuracy, able to
generatively recover the level and payoff but not the switch rate.

\begin{longtable}[]{@{}
  >{\raggedright\arraybackslash}p{(\linewidth - 14\tabcolsep) * \real{0.1525}}
  >{\raggedright\arraybackslash}p{(\linewidth - 14\tabcolsep) * \real{0.1102}}
  >{\raggedright\arraybackslash}p{(\linewidth - 14\tabcolsep) * \real{0.0678}}
  >{\raggedright\arraybackslash}p{(\linewidth - 14\tabcolsep) * \real{0.1610}}
  >{\raggedright\arraybackslash}p{(\linewidth - 14\tabcolsep) * \real{0.1186}}
  >{\raggedright\arraybackslash}p{(\linewidth - 14\tabcolsep) * \real{0.1610}}
  >{\raggedright\arraybackslash}p{(\linewidth - 14\tabcolsep) * \real{0.0678}}
  >{\raggedright\arraybackslash}p{(\linewidth - 14\tabcolsep) * \real{0.1610}}@{}}
\caption{\textbf{Generative fit of the winning models for the inspection
game.} \emph{obs} is the observed statistic and \emph{gen} is the
generated posterior-predictive mean with its 95\% predictive interval in
brackets. Bold marks statistics accepted under the ROPE equivalence test
where more than 97.5\% of the discrepancy distribution fell within
±0.05. Gemini under normal prompting had no learning model that beat the
mixed-equilibrium reference and was therefore not
included.}\label{tbl-ig-ppc}\tabularnewline
\toprule\noalign{}
\begin{minipage}[b]{\linewidth}\raggedright
Group
\end{minipage} & \begin{minipage}[b]{\linewidth}\raggedright
Winning model
\end{minipage} & \begin{minipage}[b]{\linewidth}\raggedright
p(Work) obs
\end{minipage} & \begin{minipage}[b]{\linewidth}\raggedright
p(Work) gen
\end{minipage} & \begin{minipage}[b]{\linewidth}\raggedright
p(Switch) obs
\end{minipage} & \begin{minipage}[b]{\linewidth}\raggedright
p(Switch) gen
\end{minipage} & \begin{minipage}[b]{\linewidth}\raggedright
Payoff obs
\end{minipage} & \begin{minipage}[b]{\linewidth}\raggedright
Payoff gen
\end{minipage} \\
\midrule\noalign{}
\endfirsthead
\toprule\noalign{}
\begin{minipage}[b]{\linewidth}\raggedright
Group
\end{minipage} & \begin{minipage}[b]{\linewidth}\raggedright
Winning model
\end{minipage} & \begin{minipage}[b]{\linewidth}\raggedright
p(Work) obs
\end{minipage} & \begin{minipage}[b]{\linewidth}\raggedright
p(Work) gen
\end{minipage} & \begin{minipage}[b]{\linewidth}\raggedright
p(Switch) obs
\end{minipage} & \begin{minipage}[b]{\linewidth}\raggedright
p(Switch) gen
\end{minipage} & \begin{minipage}[b]{\linewidth}\raggedright
Payoff obs
\end{minipage} & \begin{minipage}[b]{\linewidth}\raggedright
Payoff gen
\end{minipage} \\
\midrule\noalign{}
\endhead
\bottomrule\noalign{}
\endlastfoot
Human & Influence play & \textbf{0.336} & \textbf{0.352} {\scriptsize
{[}.34, .36{]}} & \textbf{0.455} & \textbf{0.479} {\scriptsize {[}.47,
.49{]}} & \textbf{0.545} & \textbf{0.524} {\scriptsize {[}.51,
.53{]}} \\
DeepSeek & RL & 0.014 & 0.662 {\scriptsize {[}.58, .74{]}} & 0.004 &
0.066 {\scriptsize {[}.05, .08{]}} & 0.359 & 0.542 {\scriptsize {[}.52,
.56{]}} \\
DeepSeek-SCoT & Fictitious play & 0.396 & 0.342 {\scriptsize {[}.33,
.35{]}} & \textbf{0.393} & \textbf{0.396} {\scriptsize {[}.39, .41{]}} &
\textbf{0.502} & \textbf{0.530} {\scriptsize {[}.52, .54{]}} \\
Gemini-SCoT & Influence play & \textbf{0.234} & \textbf{0.261}
{\scriptsize {[}.25, .27{]}} & 0.223 & 0.292 {\scriptsize {[}.28,
.30{]}} & \textbf{0.448} & \textbf{0.465} {\scriptsize {[}.46,
.47{]}} \\
GPT-4.1 & RL & 0.048 & 0.591 {\scriptsize {[}.49, .69{]}} & 0.044 &
0.091 {\scriptsize {[}.07, .11{]}} & 0.371 & 0.518 {\scriptsize {[}.49,
.54{]}} \\
GPT-4.1-SCoT & Influence play & \textbf{0.466} & \textbf{0.436}
{\scriptsize {[}.43, .45{]}} & \textbf{0.696} & \textbf{0.687}
{\scriptsize {[}.67, .70{]}} & \textbf{0.556} & \textbf{0.521}
{\scriptsize {[}.51, .53{]}} \\
GPT-5 & Influence play & \textbf{0.343} & \textbf{0.324} {\scriptsize
{[}.32, .33{]}} & 0.531 & 0.426 {\scriptsize {[}.42, .44{]}} &
\textbf{0.533} & \textbf{0.516} {\scriptsize {[}.51, .52{]}} \\
\end{longtable}

\begin{figure}

\centering{

\includegraphics[width=0.6\linewidth,height=\textheight,keepaspectratio]{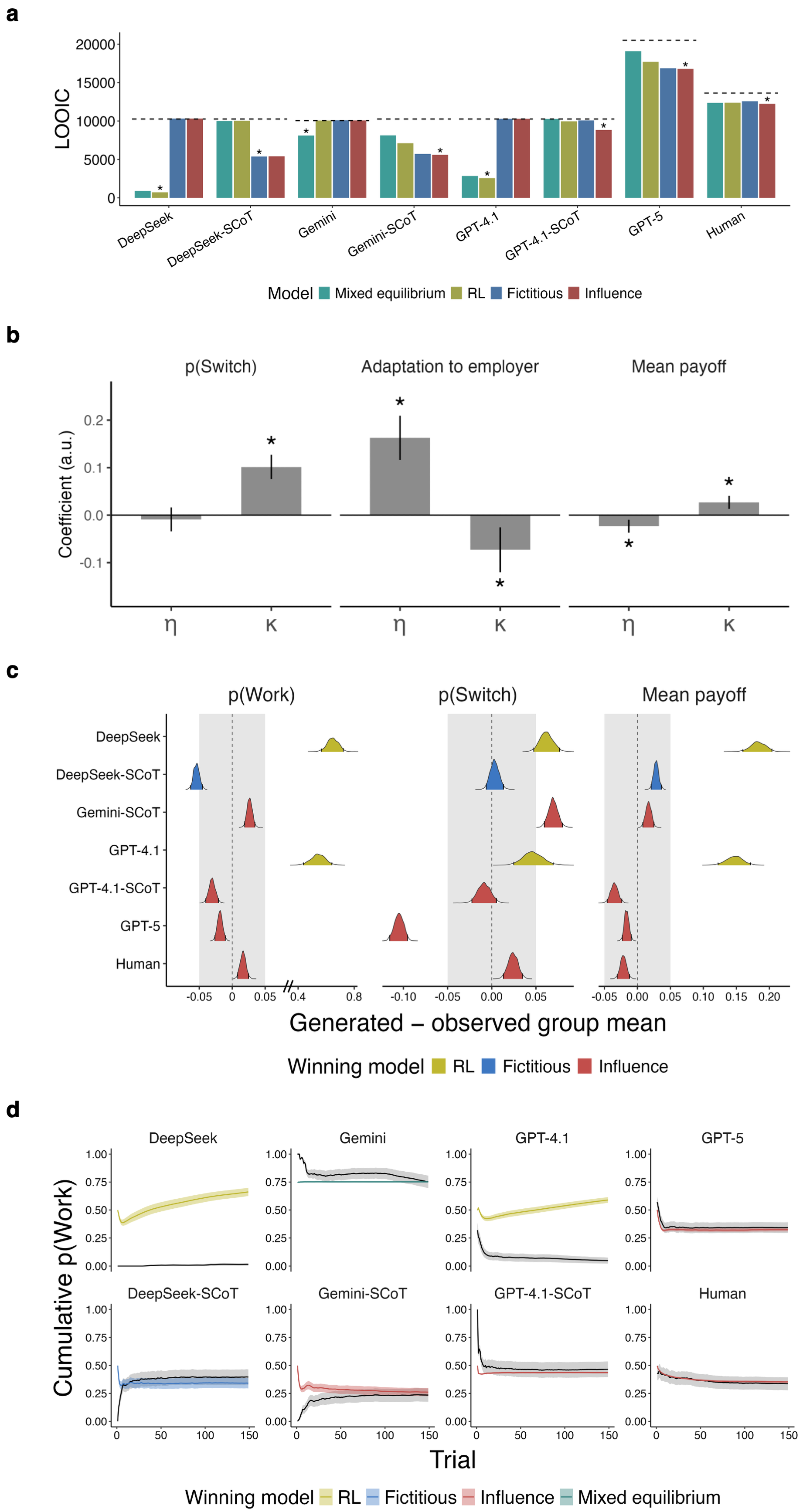}

}

\caption{\label{fig-trial-level}\textbf{Predictive and generative fit
for the inspection game.} \textbf{(a)} Model comparison across groups
using the leave-one-out information criterion (LOOIC); winning models
are highlighted. The dashed line above each group indicates the LOOIC
from chance-level responding. \textbf{(b)} Standardised coefficients
from the within-group regression of each winning model's parameters on
choice behaviour from human participants. \textbf{(c)}
Posterior-predictive discrepancy (generated minus observed group mean)
for each group's winning model, with the full ROPE-shaded discrepancy
distribution. \textbf{(d)} The rolling probability of working for the
generated against the observed data.}

\end{figure}%

Taken together, human strategic learning in the inspection game was
robustly well-explained by second-order mentalizing, where the influence
play model displayed strong predictive and generative accuracy, and its
parameters mapped strongly onto behaviour. Amongst LLM groups, Social
Chain-of-Thought prompting robustly moved LLMs towards strategic
mentalization from model-free reinforcement learning or non-learning.
These models were generatively adequate for strategic mentalization and
inadequate for RL, with the latter reflecting rigid patterns of choice
behaviour.

\subsection{Assessing adaptive mentalization using
rock-paper-scissors}\label{sec-results-rps}

To assess \emph{adaptive} mentalization, we firstly examined whether the
mean payoff differed across humans (n = 172 retained for behavioural
analyses) and LLMs (all n \textgreater{} 187) in the RPS game, when
playing against algorithms programmed to reason at different k levels (k
= 0, 1, or 2) (Fig. 4a). Separating groups by prompting strategy and
reasoning level of the opponent, we ran a linear mixed-effects model
with random intercepts for participants, given the unbalanced design
where humans played against all three bot levels while LLM participants
played against a single bot level. LLMs consisted of three groups:
DeepSeek, DeepSeek-SCoT, and GPT-5. Two models included in the previous
task (Gemini-Flash-2.0 and GPT-4.1) were not included due to technical
issues and monetary constraints respectively.

Doing so revealed significant main effects of group
(F\textsubscript{(3,~1363.2)} = 768.3, p =
2.31×10\textsuperscript{-292}, η²\textsubscript{p} = 0.63) and bot level
(F\textsubscript{(2,~2096.8)} = 439.8, p =
3.25×10\textsuperscript{-160}, η²\textsubscript{p} = 0.30), as well as a
significant interaction (F\textsubscript{(6,~1908.3)} = 306.3, p =
3.5×10\textsuperscript{-275}, η²\textsubscript{p} = 0.49). Overall,
GPT-5 performed best, earning significantly higher payoffs than humans
(t\textsubscript{(661)} = 25.5, p \textless{} 1×10\textsuperscript{-10},
d = 1.58, 95\% CI {[}1.45, 1.71{]}), and both DeepSeek variants
(vs.~DeepSeek: t\textsubscript{(2228)} = 40.7, p \textless{}
1×10\textsuperscript{-10}, d = 2.44, 95\% CI {[}2.32, 2.56{]};
vs.~DeepSeek-SCoT: t\textsubscript{(2228)} = 42.2, p \textless{}
1×10\textsuperscript{-10}, d = 2.53, 95\% CI {[}2.40, 2.65{]}). GPT-5
also outperformed all other groups at each bot level (all p \textless{}
0.001). Conversely, both DeepSeek (t\textsubscript{(661)} = 14.0, p
\textless{} 1×10\textsuperscript{-10}, d = 0.86, 95\% CI {[}0.74,
0.99{]}) and DeepSeek-SCoT (t\textsubscript{(659)} = 15.4, p \textless{}
1×10\textsuperscript{-10}, d = 0.95, 95\% CI {[}0.83, 1.07{]}) earned
significantly less than humans. The two DeepSeek groups also did not
differ overall (t\textsubscript{(2228)} = 1.44, p = 0.477, d = 0.09,
95\% CI {[}-0.03, 0.20{]}).

Performance also strongly differed across groups in response to
increasing opponent sophistication. Whilst humans showed a modest
performance decline with increasing sophistication (β = -0.032, SE =
0.007, t\textsubscript{(686)} = -4.45, p = 9.96×10\textsuperscript{-6},
d = -0.24), DeepSeek performed significantly worse (β = -0.233, SE =
0.007, t\textsubscript{(2224)} = -33.22, p =
7.73×10\textsuperscript{-197}, d = -1.73), whilst DeepSeek-SCoT also
performed significantly worse but less severely (β = -0.116, SE = 0.007,
t\textsubscript{(2224)} = -16.53, p = 5.29×10\textsuperscript{-58}, d =
-0.86). In contrast, GPT-5 was the only group to show improved
performance with increasing bot sophistication (β = 0.057, SE = 0.007,
t\textsubscript{(2224)} = 8.11, p = 8.4×10\textsuperscript{-16}, d =
0.42), a pattern significantly different from all other groups (all p
\textless{} 0.001). SCoT prompting did not significantly change
DeepSeek's overall payoff (p = 0.477), but attenuated its performance
decline against more sophisticated opponents (difference in slopes:
t\textsubscript{(2224)} = 11.82, p \textless{} 0.001, d = 0.87).

To uncover latent unobserved mechanisms of adaptive mentalization, we
fit computational models representing specific decision-making
strategies. This included the `Cognitive Hierarchy Assessment' (CHASE)
model, previously developed and validated among human participants to
infer moment-by-moment adaptive changes in mentalization strategy from
observed choice behaviour {[}\citeproc{ref-buergi2026}{107}{]}. The
CHASE model dynamically tracks a player's changing belief about the
level of cognitive sophistication of the opponent, determining the
appropriate mentalization strategy on each trial. Additional
computational models representing less sophisticated and fixed
strategies were also fitted separately across all groups (see Methods).
Conversely to humans (n = 184 retained for modeling) who played against
all three opponent levels in sequence, LLMs played separate games of a
single block against each k-level opponent. To match the existing
protocol and to facilitate efficient parameter estimation, individual
LLM instances were combined to create LLM participants consisting 3
blocks of 40 trials each, representing each programmed bot-level
\emph{k}, where \emph{k} ∈ \{0, 1, 2\}. This approach is validated
theoretically by the models - which do not model memory effects across
blocks - and practically by a permutation analysis across individual LLM
instances, revealing no significant differences in parameter estimation
(see Supplementary Information section 2). To determine the winning
model, we applied Random effects Bayesian model selection (RFX-BMS)
{[}\citeproc{ref-rigoux2014}{121}{]} which revealed different winning
models across groups. Specifically, for DeepSeek, CHASE was the clear
winner (PXP = 1.00), substantially outperforming all alternatives. In
contrast, DeepSeek with the SCoT prompting condition was best
characterized by Fictitious play (PXP = 0.99), a model where agents
estimate the probabilities with which the other player chooses their
actions and then best respond to them
{[}\citeproc{ref-hampton2008}{80}{]}. On the other hand, GPT-5 showed a
unanimous preference for the Theory-of-Mind k (ToMk) model
{[}\citeproc{ref-weerd2018}{122}{]} (PXP = 1.00), which assumes that
individuals reason about what their opponent is likely to do and may
consider multiple layers of strategic thinking. Finally, human
participants were best described by CHASE (PXP = 1.00), consistent with
previous results. Good-to-excellent parameter recovery further validated
the winning model for each group (Supplementary Figure 4). Notably,
CHASE provided the absolute best fit across all datasets, as measured by
the negative log-likelihood, but was penalised for model complexity (See
Appendix for exact model fits).

To assess the degree to which LLMs and human players were able to
correctly adapt their behaviour to the perceived recursive level of the
opponent, we examined belief distributions from the CHASE model,
representing the trial-by-trial estimate of the probability that the
opponent is operating at each reasoning level, for each subject
individually (see Methods). Although ToMk provided the best overall fit
for GPT-5, CHASE fitted choices equally well in raw log-likelihood
(paired t\textsubscript{(187)} = 0.70, p = 0.48). We therefore used
CHASE as a common space for interpreting trial-level reasoning, a
mechanism not possible among other models. When doing so, we
specifically restricted trials to only those where beliefs exceeded a
threshold of 0.5, reflecting a dominant belief, following similar
analyses performed previously {[}\citeproc{ref-buergi2026}{107}{]} (Fig.
4b). Of note, doing so removed almost half of trials (45.4\%) where no
dominant belief was expressed. DeepSeek-SCoT agents were determined to
not form adaptive beliefs when fitted with the CHASE model, with
participants almost exclusively estimated to operate at κ\textless2, and
were therefore not suitable for analysis.

To formally test whether accuracy differed jointly by model and opponent
sophistication, we ran a mixed-design repeated-measures ANOVA on each
subject's percentage of correctly-inferred trials, where subjects
missing data for any opponent level were excluded (DeepSeek n=97, GPT-5
n=150, Human n=126). As Mauchly's test indicated a violation of
sphericity, Greenhouse-Geisser-corrected p-values were calculated for
within-subject effects. Doing so revealed significant main effects of
model (F\textsubscript{(2,370)} = 224.91, p =
1.20×10\textsuperscript{-64}) and opponent level
(F\textsubscript{(2,740)} = 251.68, p = 2.10×10\textsuperscript{-77}, ε
= 0.917), and a significant model × opponent-level interaction
(F\textsubscript{(4,740)} = 64.55, p = 3.51×10\textsuperscript{-43}),
confirming that the effect of opponent sophistication on adaptive
mentalization differed by group.

Next, we sought to determine whether agents' were correctly able to
adapt to the opponent, using a repeated-measures Friedman test across
the four k-levels for each group and opponent-level combination. Where
this test was significant, we subsequently ran one-sided Wilcoxon
signed-rank tests, with an a priori hypothesis for a higher percentage
of trials for the k-level one higher than the opponent, indicating
adaptive mentalization. Each class of tests were Holm-corrected
(α=0.05). Friedman tests firstly confirmed non-uniform belief
distributions across the four candidate levels (all
p\textsubscript{holm} \textless{} 0.001). Signed-rank tests subsequently
revealed that GPT-5 agents adapted appropriately across all opponent
levels, with trials for k+1 significantly higher than every alternative
level (all p\textsubscript{holm} ≤ 8.3×10\textsuperscript{-21}). For
DeepSeek, this held against the simplest opponent (k=0; all three
comparisons p\textsubscript{holm} \textless{}
5.3×10\textsuperscript{-18}) and partially against the intermediate
opponent (k=1; correct level k=2 only exceeded k=0 and k=3, both
p\textsubscript{holm} = 1.81×10\textsuperscript{-9}; but not k=1,
p\textsubscript{holm} ≈ 1), but not the most sophisticated opponent
(k=2; all three p\textsubscript{holm} = 1.0). Humans showed the same
pattern with strong adaptation towards k=0 (all three
p\textsubscript{holm} ≤ 1.1×10\textsuperscript{-3}), partial adaptation
against k=1 (correct level k=2 exceeded k=3, p\textsubscript{holm} =
1.74×10\textsuperscript{-14}, but not k=0 or k=1, both
p\textsubscript{holm} = 0.339), but not against the most sophisticated
opponent (k=2; all p\textsubscript{holm} ≈ 1.0).

To relate adaptation to performance, we then correlated each subject's
proportion of trials where they adapted correctly with their mean score
in the task for each group. We specifically restricted these analyses to
agents deemed to adapt to the opponent (i.e., an estimated
κ\textgreater1; DeepSeek n=139, GPT-5 n=182, Human n=130). Following the
prior analyses, we similarly defined `correct' trials where a definitive
belief (\textgreater{} 0.5) was held at the adaptive level (Fig. 4c). We
also performed the same tests using the single highest belief, as
excluded trials were incorrect by definition in the former method,
resulting in many agents with 0\% accuracy. We used Spearman-rank
correlations with Holm-correction across the 6 tests. Although none of
the correlations survived correction, the trend was generally positive
across groups and analyses. Restricting trials to dominant beliefs
revealed a positive trend for DeepSeek (Spearman ρ = 0.180,
p\textsubscript{adj} = 0.2032), humans (Spearman ρ = 0.163,
p\textsubscript{adj} = 0.2567), and GPT-5 (Spearman ρ = 0.120,
p\textsubscript{adj} = 0.3171). Similarly, using the winner-take-all
approach revealed similar positive trends for DeepSeek (ρ = 0.119,
p\textsubscript{adj} = 0.3257) and humans (ρ = 0.175,
p\textsubscript{adj} = 0.2331) but a slight negative trend for GPT-5 (ρ
= −0.073, p\textsubscript{adj} = 0.3279). Altogether, these analyses
highlight distinct patterns of adaptive mentalization across groups,
which positively corresponds with improved task performance, albeit
weakly.

\begin{figure}

\centering{

\includegraphics[width=1\linewidth,height=\textheight,keepaspectratio]{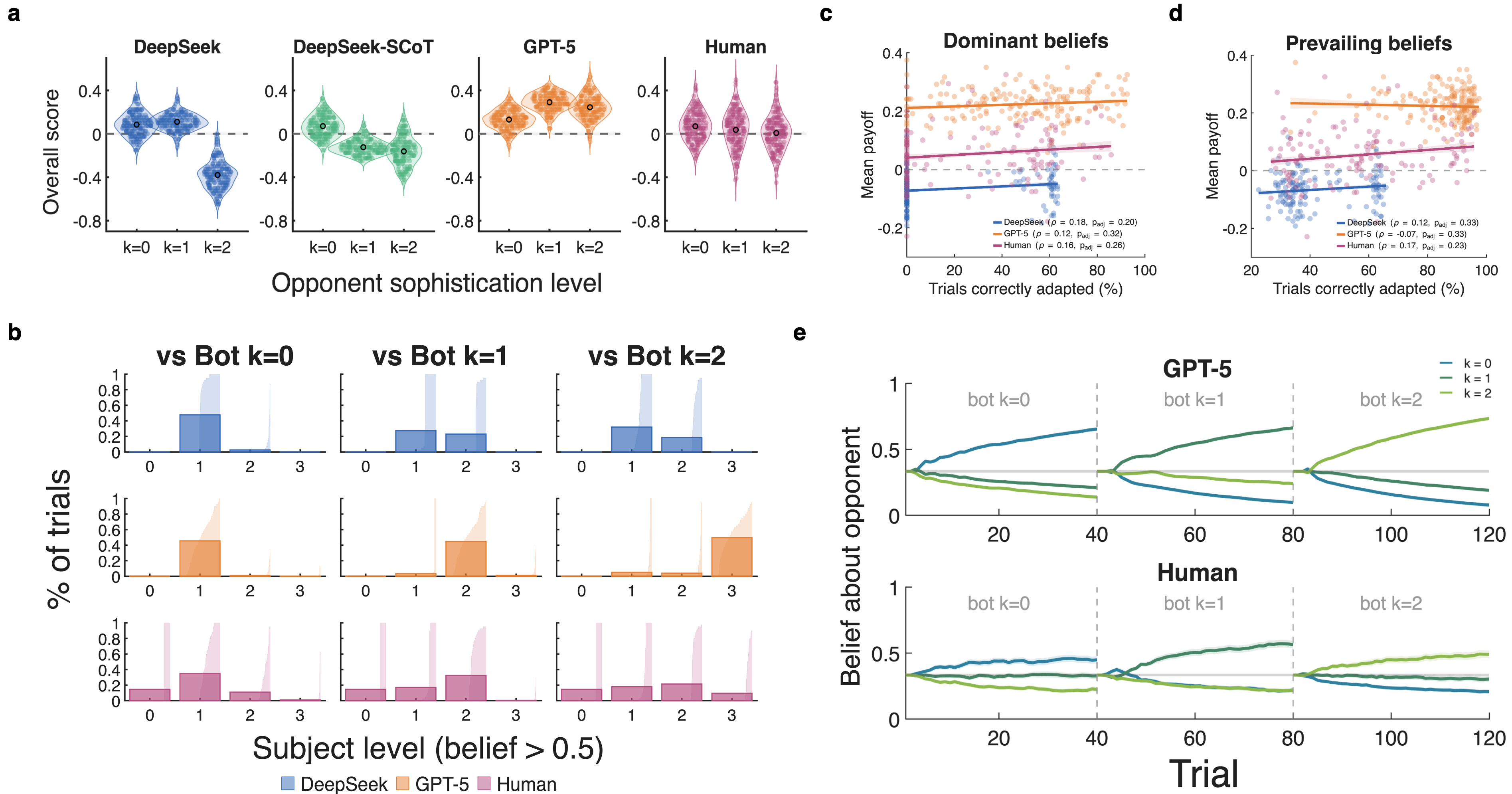}

}

\caption{\label{fig-rps}\textbf{Signatures of adaptive mentalization in
humans and large language models.} \textbf{(a)} Behavioural performance
on the task, determined by the mean score, split by group, revealed
differences between humans and LLMs. Whilst both DeepSeek groups and
humans perform worse with increasing bot sophistication, GPT-5
conversely improves. \textbf{(b)} The inferred level of mentalization
among agents plotted as a proportion of trials per subject in each
group. Reflecting the behavioural differences in performance, DeepSeek
and human participants adapt well to low-adaptive reasoners, but
struggle with opponents capable of higher-order mentalization.
Conversely, GPT-5 participants were able to adapt well to all reasoning
levels. For visualization, prevailing beliefs (\textgreater0.5) were
taken directly from the CHASE model and adjusted to reflect the k-level
of the agent by adjusting one level higher. DeepSeek-SCoT participants
were excluded from the plot, being almost exclusively estimated to have
a value of k\textless2 when fit with the CHASE model. \textbf{(c)}
Correlations between the percentage of trials where agents correctly
adapted to the inferred mentalising level of the opponent, defined using
a dominant belief. \textbf{(d)} Correlations between the percentage of
trials where agents correctly adapted to the inferred mentalising level
of the opponent, defined using a prevailing belief. \textbf{(e)}
CHASE-inferred beliefs about opponent sophistication across play. For
each agent group (top: GPT-5, n = 174; bottom: humans, n = 80), lines
show the model's mean posterior belief that the opponent is reasoning at
each level across the 120 trials (κ = 3 participants only). A grey
horizontal line denotes chance level (1/3); shaded bands are ±1 SEM.}

\end{figure}%

Finally, we analysed whether agents could correctly infer their
opponent's sophistication by examining the CHASE model's posterior
belief in the true opponent level on the final trial of each block. As
less than 2\% of DeepSeek agents met this criteria, we only conducted
these analyses where κ=3 was the estimated level, across GPT-5 (n = 174)
and human participants (n = 80) (Fig. 4e), where human participants
missing a final belief for k=0 (two missing) and k=1 (one missing)
opponents and were excluded. One sample, two-tailed t-tests against
chance (1/3) revealed that both GPT-5 (k=0: t\textsubscript{(173)} =
22.30, p = 9.18×10\textsuperscript{-53}, d = 1.69; k=1:
t\textsubscript{(173)} = 23.38, p = 1.99×10\textsuperscript{-55}, d =
1.77; k=2: t\textsubscript{(173)} = 29.72, p =
7.3×10\textsuperscript{-70}, d = 2.25) and humans (k=0:
t\textsubscript{(77)} = 4.23, p = 6.45×10\textsuperscript{-5}, d = 0.48;
k=1: t\textsubscript{(78)} = 7.86, p = 1.74×10\textsuperscript{-11}, d =
0.88; k=2: t\textsubscript{(79)} = 5.35, p =
8.52×10\textsuperscript{-7}, d = 0.60) held above-chance correct beliefs
at every sophistication level. A linear mixed model (belief
\textasciitilde{} agent × level + (1 \textbar{} subject)) revealed a
significant agent × opponent interaction (F\textsubscript{(2,~505.0)} =
19.79, p = 5.33×10\textsuperscript{-9}, η²\textsubscript{p} = 0.07) with
main effects for agent (F\textsubscript{(1,~418.7)} = 48.38, p =
1.36×10\textsuperscript{-11}, η²\textsubscript{p} = 0.10) and level
(F\textsubscript{(2,~504.2)} = 22.30, p = 5.22×10\textsuperscript{-10},
η²\textsubscript{p} = 0.08). This highlights the trend where GPT-5's
final belief also increased with opponent depth, whereas humans showed a
weaker, inverted-U pattern peaking against the k=1 opponent. Comparing
across both groups, GPT-5's beliefs were significantly more accurate
than humans' at every level (Welch's t; k=0: t\textsubscript{(121.2)} =
6.63, p = 1×10\textsuperscript{-9}, d = 0.99; k=1:
t\textsubscript{(115.5)} = 3.05, p = 0.003, d = 0.47; k=2:
t\textsubscript{(113.9)} = 7.60, p = 9.14×10\textsuperscript{-12}, d =
1.18), with the advantage widest against the most sophisticated (k=2)
opponent. All comparisons survived Bonferroni correction, with 6
comparisons (α = 0.0083) for tests against chance, and 3 comparisons (α
= 0.0167) for group comparisons. Therefore, while both humans and GPT-5
tracked opponent sophistication above chance, GPT-5 did so more
accurately and, unlike humans, increasingly so as opponents grew more
sophisticated.

\section{Discussion}\label{sec-discussion}

A growing interest concerns whether artificial intelligence systems
understand the thoughts and beliefs of others, and use this information
to guide their own actions. Previous studies have assessed
theory-of-mind in LLMs by measuring choice accuracy in response to
story-based prompts, obscuring the latent mechanisms underlying observed
behaviour. Here we employ two behavioural tasks previously validated in
human participants, the inspection game and rock-paper-scissors (RPS).
These tasks provide a normative assessment of machine intelligence
{[}\citeproc{ref-hu2025}{65}{]}, and allow the examination of behaviour
at specific depths of mentalizing {[}\citeproc{ref-wagner2025}{69}{]}.

Across two strategic interaction games, humans demonstrated both
recursive and adaptive mentalization, replicating previous results and
demonstrating the robustness of the tasks used
{[}\citeproc{ref-hampton2008}{80}, \citeproc{ref-hill2017}{106},
\citeproc{ref-buergi2026}{107}{]}. On the other hand, LLMs demonstrated
varying mentalization strategies across the two strategic games, with
significant differences observed across model providers. Furthermore,
LLMs prompted using Social Chain-of-Thought (SCoT) demonstrated robust
improvements in performance, reflecting more sophisticated reasoning.
Yet, groups differed in their generative accuracy, limiting
interpretations towards latent processes of mentalization. In the RPS
game, GPT-5 agents demonstrated efficient adaptive mentalization, whilst
DeepSeek agents appropriately adapted to bots playing at the lowest
level but struggled to adapt to complex reasoners. Compared to prior
studies assessing mentalization using story-based one-shot tasks or
binary choice paradigms, our approach leverages trial-level
computational modeling to determine the specific latent mechanisms
underlying mentalization. These models can be subject to rigorous
validation by determining predictive and generative accuracy, allowing
for the direct mapping of mechanism to behavior.

Cognitive computational modeling, as implemented in our study, offers a
normative tool for quantifying latent (social) cognitive processes and
tracking how they change in response to targeted interventions
{[}\citeproc{ref-zhang2020a}{79},
\citeproc{ref-lockwood2021}{123}--\citeproc{ref-sohail2024}{126}{]}.
Previous work has examined differences in mentalization across several
species of non-human primates {[}\citeproc{ref-devaine2017}{14}{]}. We
use these principles to determine whether similar behavioural outcomes
across LLMs and humans reflect shared latent computations of
mentalization. Our results demonstrate that this framework is also
sufficiently sensitive to detect meaningful differences in mentalizing
behaviour across model providers, and captures the effect of strategic
prompting, a manipulation comparable to language-based psychotherapy in
humans {[}\citeproc{ref-ben-zion2025}{127}{]}. Prompting strategies
designed to improve ToM in LLMs should ultimately be evaluated against
process-level measures, as relying on surface-level patterns may obscure
latent changes in social cognition {[}\citeproc{ref-hu2025}{65},
\citeproc{ref-lu2025}{70}{]}. Yet, whilst providing a deeper level of
understanding than assessing choices, computational modeling still
obscures the mechanistic components of LLM which must instead be
inferred from their observed behaviour
{[}\citeproc{ref-palminteri2025}{110}{]}. This is further mired by the
inferential properties of LLMs, which as pseudo-statistical machines
utilize traditional stochastic processes and human-like abductive
reasoning to reach conclusions {[}\citeproc{ref-floridi2025}{128},
\citeproc{ref-yetman2025}{129}{]}. Despite the emergence of frameworks
and best practices for determining computational equivalence between
humans and LLMs {[}\citeproc{ref-frank2025}{23},
\citeproc{ref-ku2025}{66}, \citeproc{ref-connell2024}{72},
\citeproc{ref-palminteri2025}{110}, \citeproc{ref-beckmann2025}{130},
\citeproc{ref-butlin2025}{131}{]}, doing so currently remains a
challenging task. This is evidenced by the opaque definition of
theory-of-mind in machines, which requires an interdisciplinary
approach, one incorporating theories from cognitive psychology,
neuroscience and artificial intelligence
{[}\citeproc{ref-cuzzolin2026}{132}{]}. Looking ahead, a comprehensive
assessment of mentalization in LLMs will likely require a multi-modal
approach involving multiple levels of explanation
{[}\citeproc{ref-frank2025}{23}, \citeproc{ref-ivanova2025}{133},
\citeproc{ref-wang2026}{134}{]}.

The observed differences in task performance may be partly attributed to
the model size, where GPT models may constitute significantly more
parameters and are trained upon larger quantities of data than the
DeepSeek and Gemini models used in the tasks. Large language models are
based on the transformer architecture of neural networks, pretrained on
a large text-based corpus. Subsequently, the neural scaling law
demonstrates that their performance generally improves as the model size
and training data increase {[}\citeproc{ref-kaplan2020}{135}{]}. Larger
transformer models also demonstrate emerging properties absent in
smaller models {[}\citeproc{ref-berti2025}{136}{]} including in-context
learning {[}\citeproc{ref-wang2025}{137}{]} and multi-step reasoning
{[}\citeproc{ref-ruan2024}{138}{]}. Subsequently, model size in LLMs has
been shown to robustly correlate with performance across benchmarks
assessing economic and rational choice
{[}\citeproc{ref-coda-forno2024}{112},
\citeproc{ref-mina2025}{139}--\citeproc{ref-zhou2025}{142}{]}. Larger
models often may possess longer context histories, capable of accurately
retrieving and processing more information when making choices. This is
supported by comprehension checks for the tasks used in the current
study which report differences with the retrieval accuracy of
task-related information across model sizes (see Supplementary
Information section 3).

Modeling analyses from the RPS task revealed DeepSeek was unable to
appropriately calibrate its recursive level of thinking to the opponent,
an ability not significantly improved by SCoT prompting. On the other
hand, GPT-5 agents were able to accurately form beliefs across all
recursive levels. These results follow recent evidence suggesting that
smaller LLMs generally are able to accurately infer others' perceptions
{[}\citeproc{ref-jung2024}{143}{]}, but struggle to appropriately form
and update their beliefs about other agents
{[}\citeproc{ref-imran2025}{144}{]}. Furthermore, whilst LLMs are
generally inconsistent in how they update their beliefs
{[}\citeproc{ref-pal2025}{145}{]}, larger models learn to update their
beliefs about propositions more consistently
{[}\citeproc{ref-imran2025}{144}{]}. Unlike human-based belief updating
which is bounded by data and computational constraints
{[}\citeproc{ref-zhu2026}{146}{]}, LLMs are able to burden more
computationally intensive processes of reasoning over extended periods.
Taken together, GPT-5 agents may be able to more accurately predict the
actions of complex reasoners by using stored choice histories of the
opponent together with deliberative reasoning. Strategic prompting is
widely used for evoking complex, human-like cognitive operations among
LLMs {[}\citeproc{ref-wang2024}{47},
\citeproc{ref-ebouky2025}{147}--\citeproc{ref-patil2025}{150}{]}
including theory-of-mind (ToM)
{[}\citeproc{ref-chen2025b}{151}--\citeproc{ref-wilf2023}{155}{]}, where
performance is improved by appropriate prompting which elevate ToM
reasoning {[}\citeproc{ref-moghaddam2023}{153}{]}. However,
prompting-induced improvements in mentalization are context-dependent
{[}\citeproc{ref-moghaddam2023}{153}{]} and task performance can more
generally be improved by providing strategy-specific information within
prompts {[}\citeproc{ref-mondorf2024}{67},
\citeproc{ref-zhang2025}{156}{]}. This altogether suggests that the SCoT
prompt may be inappropriately vague for smaller models given the
specific task demands of the RPS game. Prompts including additional task
or strategy-specific information could potentially elicit adaptive
mentalization in models less capable of sophisticated reasoning.

We note that a mechanistic explanation of mentalization is ultimately
not possible due to the closed-sourced nature of the models used.
However, specific activity patterns of hidden embeddings across several
open-sourced LLMs are found to correspond with their ability to
represent another's perspective, where the percentage of embeddings
displaying significant selectivity positively correlates with model size
{[}\citeproc{ref-jamali2023}{157}{]}. Furthermore, recent mechanistic
work has suggested that social reasoning - including ToM - is
specifically encoded in extremely sparse, localised parameter patterns
tied to contextual processing {[}\citeproc{ref-wu2025}{158}{]}. An open
question therefore remains whether larger closed-source models also
demonstrate specific patterns of activity corresponding to mentalization
depth, and whether parameter-level signatures of mentalization exist
within them.

Our study has theoretical and practical implications across the fields
of artificial intelligence and machine psychology. As LLMs are
increasingly employed in different sectors, their ability to accurately
interpret and appropriately respond to human mental states becomes
critical {[}\citeproc{ref-amirizaniani2025}{159}{]}. Large language
models as chatbots are widely used as therapeutic tools
{[}\citeproc{ref-casu2024}{160}--\citeproc{ref-yuan2025}{164}{]},
capable of building artificial relationships
{[}\citeproc{ref-chu2025}{165}--\citeproc{ref-jones2025}{167}{]} and
providing emotional support
{[}\citeproc{ref-folk2025}{168}--\citeproc{ref-zhang2025a}{171}{]},
roles necessitating accurate mentalization of the user's preferences,
feelings and potential actions. Yet, LLMs often provide inappropriate or
harmful responses, a consequence of not accurately interpreting subtle
text-based cues {[}\citeproc{ref-adrian2025}{172}{]}. Indeed, AI systems
are often criticized for not being able to demonstrate affective and
motivational empathy, mentalization-derived components essential to
human-based therapy
{[}\citeproc{ref-gabriels2026}{173}--\citeproc{ref-yirmiya2025}{175}{]}.
The capacity for LLMs to mentalize also influences user perception and
interactions. Humans frequently anthropomorphise AI tools, attributing
mental states based on their own expectations and the agents' observed
behaviour
{[}\citeproc{ref-cohn2024}{176}--\citeproc{ref-inie2024}{178}{]}, which
in turn predicts trust and social closeness towards these systems
{[}\citeproc{ref-folk2025}{168}, \citeproc{ref-colombatto2025}{179}{]}.
Experimentally portraying LLMs intentionally as social entities has also
been shown to increase attributions of warmth, empathy, and competence
among users {[}\citeproc{ref-yao2025}{180}{]}, traits which can be
enhanced by inducing a theory-of-mind in LLMs
{[}\citeproc{ref-schlesener2025}{181}{]}. Finally, instilling the
capacity for complex `human-like' cognition among LLMs can also augment
and complement human thinking and decision-making
{[}\citeproc{ref-gonzalez2025}{182}{]}. Human intelligence strongly
differs from machine intelligence, where humans build causal models of
the world, and efficiently apply relevant prior learned information and
knowledge to new tasks and situations {[}\citeproc{ref-lake2017}{26}{]}.
Conversely, generative AI systems such as LLMs are statistical inference
machines based on pattern recognition. Building machines that learn and
think more strongly like humans - including a capacity for mentalization
- can lead to more knowledgeable, reliable and trustworthy systems
{[}\citeproc{ref-collins2024}{115}{]}.

We acknowledge several practical limitations of our experiments. The
analytical choices used for LLM configuration, including model parameter
settings and prompt configurations, significantly impact their accuracy
for replicating human data
{[}\citeproc{ref-cummins2025}{183}--\citeproc{ref-loya2023}{186}{]}.
Therefore, our response patterns may not be observed using different
model providers, parameters or prompts. Additionally, the fixed choice
history of 10 trials may obscure the effect of different context windows
on choice behaviour {[}\citeproc{ref-fontana2025}{43}{]}. We also
specifically focus on a specific group of model providers and do not
include any open-source models in our analyses, a recommended practice
for understanding human cognition with LLMs
{[}\citeproc{ref-frank2023}{187}{]}. Acknowledging these points, we
ultimately follow existing guidelines for evaluating the cognitive
abilities of LLMs {[}\citeproc{ref-palminteri2025}{110},
\citeproc{ref-ivanova2025}{133}{]}, aiming to test specific hypotheses
concerning the computational form of the target process - mentalization
- and the effect of a single additional prompting scheme previously
validated. We also follow the recommended practice of `machine
validation' {[}\citeproc{ref-soubki2025}{188}{]}, where statistical
patterns and strategies uniquely exploited by machines are accounted
for. In our primary analyses, we generated the experiment text without
the use of LLMs and place models in the role of an active participant
akin to human participants. Yet, LLMs may nevertheless exploit lexical
patterns present in the experimental prompt when making choices
{[}\citeproc{ref-hu2025}{65}{]}. Subsequently, we attempt to control for
various confounds including choice order and composition in our primary
analyses and formally investigate their effect in our supplementary
analyses, reporting varying effects of these confounds by model and task
(see Supplementary Information section 3). Finally, our experimental
design does not entirely dissociate a machine's knowledge (i.e.,
`competence') from it's application (i.e., `performance')
{[}\citeproc{ref-firestone2020}{189}{]}. Whilst we aimed to similarly
limit both humans and machines by providing identical task instructions
and ensuring a thorough understanding of the task rules and objectives,
the outcome history was only provided to LLMs and not humans. We
acknowledge that a similar trial history made available to humans, or
conversely, burdening machines with the equivalent human limitations,
would strengthen our claims.

Researchers are increasingly turning to methods rooted in psychology and
cognitive science to understand the behaviour of large language models
{[}\citeproc{ref-wulff2026}{190}{]}. Yet, how these systems actively
represent others remains unclear. Our findings highlight key differences
in the ability to recursively mentalize across LLMs, evident through
observed behaviour and the underlying computations. Artificial
intelligence (AI) continues to develop at an alarming rate, with some
forecasting that digital computation may eventually supersede human
intelligence {[}\citeproc{ref-hinton2024}{191}{]}. Simultaneously,
building AI systems with advanced cognition stands to improve their
application across a range of social domains, yet increases risks toward
malicious use {[}\citeproc{ref-bengio2026}{192},
\citeproc{ref-bengio2024}{193}{]}. Within this complex, rapidly evolving
space, understanding the cognition of these systems is necessary to
appropriately scale the development of AI technologies.

\section{Methods}\label{sec-methods}

\subsection{Experimental procedure}\label{sec-methods-procedure}

\subsubsection{Humans}\label{humans}

\subparagraph{Inspection game}\label{inspection-game}

The inspection game is a repeated non-cooperative game where players may
implement strategies corresponding with different levels of recursive
thinking {[}\citeproc{ref-hampton2008}{80},
\citeproc{ref-hill2017}{106}{]}. In the game, two players assume the
roles of an ``employee'' and an ``employer'' and make one of two choices
simultaneously. On each trial, the employee chooses whether to ``work''
or ``shirk,'' and the employer chooses whether to ``inspect'' or ``not
inspect'' the employee. In our version (Figure~\ref{fig-design} a), the
participant always played the role of the employee.

Each trial began with a fixation cross for 1 second, after which the
options of ``work'' or ``shirk'' were presented. Participants had up to
4 seconds to indicate their choice by pressing the respective arrow key,
and the chosen option was highlighted with a red square for up to 2
seconds. The choice of their opponent, whether to ``inspect'' or ``not
inspect'', was then presented on the screen for 3 seconds, alongside the
outcome of that trial. The task consisted of a single block of 150
trials. Points from the task were added and converted to a bonus payment
in GBP at the end of the session.

Participants were told they were playing against a real opponent, but
were actually playing against a computer algorithm based on the
influence learning model fitted to real human data from previous data
{[}\citeproc{ref-hill2017}{106}{]}. The artificial opponent's responses
were generated by the fitted influence learning model, which would both
(1) track player choices and respond optimally to their frequency and
(2) use its own past choices to form player expectations, before (3)
introducing randomness to the choices based on these two computations
(used model parameters: the first-order belief updating η = 0.3;
second-order belief updating κ = 0.1; inverse temperature β = 1.5). We
used MATLAB 2012a (MathWorks) and the Cogent 2000 v125 toolbox
(\href{http://www.vislab.ucl.ac.uk/Cogent/}{www.vislab.ucl.ac.uk/Cogent/})
to present the task.

\subparagraph{Rock-paper-scissors}\label{rock-paper-scissors}

To assess adaptive mentalization, we employed a zero-sum
rock-paper-scissors (RPS) game. In contrast to the inspection game,
which is played against a fixed algorithm across all trials,
participants in the RPS game played against an algorithm programmed to
make choices in accordance with a specific reasoning level \emph{k},
where \emph{k} ∈ \{0, 1, 2\}. Subsequently, this experimental design is
suitable for assessing whether humans and LLMs are capable of adapting
their behaviour to the \emph{k}-level of the opponent
{[}\citeproc{ref-buergi2026}{107}, \citeproc{ref-jiang2022}{194}{]}.

Participants were provided instructions that they will be playing a
simple repeated Rock-Paper-Scissors game with three different opponents
(Figure~\ref{fig-design} b). They were instructed that on each trial,
they must make a choice between the three options, Rock (using the
`Right' arrow key), Paper (using the `Up' arrow key) and Scissors (using
the `Left' arrow key), and that both players will be making the choices
at the same time. Participants had up to 3 seconds to make their choice,
after which they received feedback on the opponent's choice as well as
the reward outcome (+1 for win, -1 for loss, 0 for tie). Participants
were told that they would play 40 trials against a specific opponent,
after which they would be matched to a new opponent. In reality,
participants played against a computer algorithm designed to make
choices corresponding to different levels of mentalization, \emph{k}
(where \emph{k} ∈ 0, 1, 2). Opponents were randomised and
counterbalanced across participants.

\subsubsection{Large language models}\label{large-language-models}

\subparagraph{Inspection game}\label{inspection-game-1}

Large language models completed a translated version of the inspection
game presented to humans. To keep the information as similar to that
presented to the human sample, roles, choice options and the payoff
matrix were left unchanged. However, the reward metric was changed from
`points' to `cents', incentivizing the LLMs to perform well
{[}\citeproc{ref-todasco2025}{195}, \citeproc{ref-wang2025a}{196}{]}.
Large language models also completed 150 trials playing the role of the
employee against the same algorithmic opponent as the employer. In all
prompting scenarios (see Supplementary Information section 1), models
were presented with the game rules, payoff, choice history, and choice
question.

\subparagraph{Rock-paper-scissors}\label{rock-paper-scissors-1}

Replicating the experimental procedure used in humans, large language
models also completed 40 trials of a repeated game with the same
algorithmic opponent programmed to make choices at a given recursive
reasoning level. However, in contrast to human participants, who played
against all 3 levels, each LLM played only a single block against a
specific level. Whilst the payoff matrix was left unchanged, choice
options were recoded to prevent potential choice bias with numbers
(i.e., `1' selected more often than `2' or `3') or the conventional
choices (`Rock', `Paper', `Scissors' / `R', `P', `S') previously
highlighted {[}\citeproc{ref-vidler2025}{116}{]}. Instead, three single
letters (`J', `Q', `Z') were selected as approximately equally uncommon
letters in the English corpora. In all prompting scenarios, models were
presented with the game rules, choice history, and choice question.

Both tasks were programmed using custom code integrating the oTree
\{otree==5.11.1\} experimental platform
{[}\citeproc{ref-chen2016a}{197}{]} with the Python package botex
\{botex==0.2.0\} {[}\citeproc{ref-edossa2024}{198}{]} and LiteLLM
\{litellm==1.73.1\} for LLM API calls. Whilst in previous studies the
choice history has included all previous trials, a rolling choice
history of 10 trials was chosen to represent an approximation of working
memory among human players, whilst accounting for time and monetary
costs with practically running the experiments. No time limits were
enforced during both tasks.

\subsection{Participants}\label{sec-methods-participants}

\subsubsection{Humans}\label{humans-1}

\subparagraph{Inspection game}\label{inspection-game-2}

Human participants were recruited through Oxford University databases,
social media, email lists, and adverts in local newspapers. A priori, we
excluded individuals who studied psychology, had a history of
neurological or psychiatric disorders, or had abnormal vision.
Participants received £10 per hour and a bonus of up to £5 based on
their task performance (including several different tasks). All
participants provided written informed consent. The Oxford University
Medical Sciences Inter Divisional Research Ethics Committee and National
Health Service Ethics approved the study. For behavioural analyses, only
those with no more than 5 missing trials were included, leaving 67
participants (25 males, mean age = 23.3 ±4.21). All of these
participants' data were also included for modeling.

\subparagraph{Rock-paper-scissors}\label{rock-paper-scissors-2}

Participants were recruited from the general population on Prolific,
restricted to UK residents and native English speakers as part of a
larger study. Before completing the RPS task, they filled out
demographic questionnaires and questions using verbal mentalization
tasks.

Participants used their own devices to access the study on Qualtrics and
Cognition and were paid £18 for an anticipated 120 minutes of work
(median time: 117 minutes). The study was approved by the University of
Birmingham STEM Ethics Committee (ERN 22-1192, ERN 09-719F, ERN 2311).

A total of 185 human participants completed the original experiment. For
behavioural analyses, only those with less than 5 missing trials were
included, leaving 172 participants (73 males, 97 females, 2 missing;
mean age = 24.95 ±3.57 years). For modeling analyses, a single
participant was excluded for not having valid parameter estimates,
leaving 184 human participants (76 males, 106 females, 2 missing; mean
age = 24.91 ±3.57 years) for all modeling analyses.

\subsubsection{Large language models}\label{large-language-models-1}

The following LLMs were selected for participation in the respective
tasks. Inspection game: Gemini, DeepSeek, GPT-4.1, GPT-5; RPS: DeepSeek,
GPT-5. These models were selected based upon time and monetary costs
when running the study, and from primary comprehension checks that
assessed a wider set of candidate models (see Supplementary Information
section 3). Given the commonly reported issues with model response
formats and content {[}\citeproc{ref-liu2024}{199}{]}, several steps
were made to ensure reliability in model outputs. Firstly, API calls
were made to model providers using LiteLLM, unifying model communication
with a common interface. Under this framework, structured outputs in
JSON format were strictly enforced across all models using the
Instructor \{instructor==v1.6.0\} Python library where not natively
supported. Secondly, strict Pydantic schemas defined the required
response structure and data types. Finally, model responses were
programmatically limited to those defined as response options within the
prompt (i.e., `WORK'/`SHIRK' or `J'/`Q'/`Z').

Following a power analysis (see Power calculations), the sample size per
LLM per experiment consisted of:

\begin{itemize}
\item
  Inspection game: GPT-4.1 (50 standard prompt, 50 SCoT), DeepSeek (50
  standard prompt, 50 SCoT), Gemini (49 standard prompt, 50 SCoT), GPT-5
  (100 Normal)
\item
  RPS (Individual LLM instances): DeepSeek (566 standard prompt, 568
  SCoT), GPT-5 (566 standard prompt)
\end{itemize}

In modeling analyses for the RPS task, individual runs were combined
within each LLM group, with remainder runs discarded, resulting in the
following sample sizes: DeepSeek (188 standard prompt, 189 SCoT), GPT-5
(188 Normal).

\paragraph{Prompting conditions}\label{prompting-conditions}

A previous study {[}\citeproc{ref-akata2025}{40}{]} demonstrated the
effect of a simple prompting strategy - `Social Chain-of-Thought' (SCoT)
- for improving LLM performance in economic games by asking the model to
form a prediction for the opponent's choice before making a decision. To
test whether this ability also extends to the current study, two
prompting conditions were tested. Firstly, in the standard prompt
condition, models were simply asked to make a choice between the two
options, with no explicit prompt for engaging in deliberate thought. In
the SCoT condition, models were additionally prompted on each trial to
predict the choice of the opponent before making their choice.

\paragraph{Parameters}\label{parameters}

A default \emph{temperature} parameter of 0 was set for Gemini,
DeepSeek, GPT-4.1 whilst a temperature of 1 was set for GPT-5 in
accordance with the restricted parameter values provided by OpenAI. A
setting of `medium' was chosen for the \emph{reasoning\_effort}
parameter with GPT-5, providing detailed reasoning without explicit
instruction, and considering monetary restrictions for the study. Token
lengths for model responses through the \emph{max\_tokens} parameter was
not limited. LLM data was collected from the periods 04/08/2025 -
17/09/2025 and 22/09/2025 - 14/10/2025 for the inspection and RPS games
respectively.

\subsection{Power calculations}\label{sec-methods-power}

\subsubsection{Inspection game}\label{inspection-game-3}

\subparagraph{Comparisons between model and human performance by
strategy}\label{comparisons-between-model-and-human-performance-by-strategy}

Comparisons between model × strategy combinations and human participants
required seven pairwise tests. Using Welch's two-tailed t-tests with
Bonferroni correction (α = 0.007) and assuming large effects (d = 0.8),
power analyses indicated approximately 36 participants per LLM group to
achieve 80\% power. This was met with our sample of at least 49 agents
per group, and an allocation ratio of 1.34 reflecting the 67
participants in the human group.

\subparagraph{Within-model task effects of strategic
prompting}\label{within-model-task-effects-of-strategic-prompting}

Following previous results, we expected to observe a significantly
higher score across all conditions in both tasks when implementing the
SCoT prompt compared to the baseline condition
{[}\citeproc{ref-akata2025}{40}{]}. This covers our analyses for
calculating group differences for choice selection, task performance and
prediction accuracy. For the inspection game, this required three
comparisons. For Welch's two-tailed t-tests with Bonferroni correction
(α = 0.017), assuming a large effect size (d = 0.8), power analyses
indicated a required sample size of 35 per LLM group to achieve 80\%
power. This was met with our sample of at least 49 agents per group, and
an allocation ratio of 1.0.

\subparagraph{Across-model task effects of strategic
prompting}\label{across-model-task-effects-of-strategic-prompting}

Differences in task performance were also expected across LLMs, across
models and prompting strategies. To justify our sample size for the 2
(prompt) × 3 (model) factorial ANOVA, we ran a power analysis for a
fixed-effects ANOVA with six groups. Expecting a medium effect size (f =
0.25, α = 0.05), we required a total sample size of N = 158 to achieve
80\% power, whilst the main effect of prompt required N = 128. Both were
exceeded by our sample of at least 49 agents per cell. For the separate
one-way comparison of the four normally-prompted models, a
fixed-effects, one-way ANOVA with four groups and a medium-to-large
effect size (0.25 \textless{} f \textless{} 0.4) required a total of 76
\textless{} N \textless{} 180 (19 \textless{} N \textless{} 45 per
group) to achieve 80\% power (α = 0.05), which was also met.

\subsubsection{Rock-paper-scissors}\label{rock-paper-scissors-3}

\subparagraph{Within-human effects of bot level on
performance}\label{within-human-effects-of-bot-level-on-performance}

To examine the effect of bot level on human performance, a one-way
repeated measures ANOVA was conducted. Expecting a medium effect size (f
= 0.25), power analyses indicated a required sample size of 28
participants to achieve 80\% power (α = 0.05) with three repeated
measurements (bot levels 0, 1, and 2). This was exceeded with our sample
of 172 human participants, which was sufficient to detect 80\% power
with a small effect size (f = 0.15).

\subparagraph{Cross-group comparisons of performance across humans and
LLMs}\label{cross-group-comparisons-of-performance-across-humans-and-llms}

Comparisons of task performance across humans and LLMs were conducted
using a linear mixed model with group and bot level as fixed effects and
subject as a random effect. For the main effect of group, expecting a
large effect size (f = 0.40), power analyses indicated a required sample
size of 13 participants per group to achieve 80\% power (α = 0.05). For
the main effect of bot level, expecting a medium effect size (f = 0.25),
power analyses indicated a total sample size of 28 participants to
achieve 80\% power. For the group × bot level interaction, expecting a
medium effect size (f = 0.25), power analyses indicated a required
sample size of 10 participants per group to achieve 80\% power. In all
tests, the calculated sample sizes were substantially exceeded with our
collected data.

\subparagraph{Within-group pairwise comparisons of bot level
effects}\label{within-group-pairwise-comparisons-of-bot-level-effects}

To examine differences in performance between bot levels within each
group, paired t-tests with Bonferroni corrections were conducted. This
required three pairwise comparisons per group (bot level 0 vs 1, 0 vs 2,
and 1 vs 2). For one-tailed paired t-tests with Bonferroni correction (α
= 0.017), assuming a medium effect size (dz = 0.50), power analyses
indicated a required sample size of 38 per group to achieve 80\% power.
This was met or exceeded across all groups with our sample sizes of 172
human participants and 566-568 participants per LLM group.

\subparagraph{Choice probability
distributions}\label{choice-probability-distributions}

To test whether choice frequencies for rock, paper, and scissors
deviated from a uniform distribution (1/3 probability for each choice),
chi-square goodness-of-fit tests were conducted for each group. For
human participants, where we expected minimal deviation from uniformity,
assuming a small effect size (w = 0.10), power analyses indicated a
required sample size of 964 trials to achieve 80\% power (α = 0.05, df =
2). For LLM groups, where we expected small to medium deviations from
uniformity, assuming a small-medium effect size (w = 0.20), power
analyses indicated a required sample size of 241 trials to achieve 80\%
power. Additionally, chi-square tests on first choice selections
required a sample size of 39 participants per group to detect a large
effect size (w = 0.50) with 80\% power. In all cases, the required
sample sizes were substantially exceeded with our collected data.

Power calculations were computed using GPower 3.1 Software
{[}\citeproc{ref-faul2009}{200}{]}. Effect sizes were labelled following
guidance {[}\citeproc{ref-cohen2013}{201}{]} suggesting values of d/dz =
0.2/0.5/0.8 (small/medium/large) for t-tests, f = 0.10/0.25/0.40 for
ANOVA effects, and w = 0.10/0.30/0.50 for chi-square tests.

\subsection{Behavioural analyses}\label{sec-methods-behavioural}

\subsubsection{Rock-paper-scissors}\label{rock-paper-scissors-4}

\subparagraph{Linear mixed model specification for task performance
(payoff)}\label{linear-mixed-model-specification-for-task-performance-payoff}

Linear mixed-effects models (LMMs) were used to assess differences in
mean payoff across model strategies and bot sophistication levels.

Two complementary LMMs were fitted to the data:

LMM1: for ANOVA, main effects, interaction effects, and simple effects
comparisons

\begin{center}

\texttt{Payoff\ \textasciitilde{}\ ModelStrategy\ *\ BotLevel\ +\ (1\textbar{}SubjectID)}

\end{center}

LMM2: for within-group comparisons (i.e., across bot-levels)

\begin{center}

\texttt{Payoff\ \textasciitilde{}\ ModelStrategy\ *\ BotLevelNumeric\ +\ (1\textbar{}SubjectID)}

\end{center}

In the models, each LLM subject was coded as per data collection, only
completing a single block. Models were subsequently fit to 172 human
participants, 566 DeepSeek participants, 568 DeepSeek-SCoT participants
and 566 GPT-5 participants. All analyses were conducted in R using the
\emph{lme4} package for model fitting, \emph{emmeans} for marginal means
and contrasts, and \emph{effectsize} for effect size calculations.

\subsection{Computational modeling}\label{sec-methods-modeling}

\subsubsection{Inspection game}\label{inspection-game-4}

\paragraph{Candidate models}\label{candidate-models}

Following prior studies {[}\citeproc{ref-hampton2008}{80},
\citeproc{ref-hill2017}{106}{]}, we fit a series of computational models
representing participants' possible strategies across the game. In the
simplest learning strategy, players may simply choose the action in the
recent past which provided the most reward through reinforcement
learning (RL). However, players may also anticipate that the opponent
will form a prediction of their choice and counter it accordingly. This
approach, termed `fictitious play', represents a strategy corresponding
to first-order recursive thinking. Players may adopt a still more
sophisticated approach by assuming the opponent adopts a first-order
approach themselves and countering this choice accordingly. Termed
`influence play', this strategy represents second-order recursive
thinking. Finally, to test whether participants optimised by removing
predictability rather than by learning, we included a mixed-equilibrium
(ME) reference model in which choices follow a Bernoulli process with a
single per-subject rate and no trial-by-trial updating
{[}\citeproc{ref-hill2017}{106}{]}. For the inspection game, the
mixed-strategy Nash equilibrium for the employee is P(Work) = 0.5.

\emph{Reinforcement learning.} The reinforcement learning (RL) model
suggests that participants do not pay attention to the opponent's
choices but simply learn from their previous actions and rewards. This
model modifies the value of each choice for each trial \emph{t} in
response to a prediction error δ weighted by a learning rate η:

\begin{equation}\phantomsection\label{eq-rl-value}{
V_{t}^{\text{Work}} = V_{t-1}^{\text{Work}} + \eta \cdot \delta_{t-1}
}\end{equation}

The prediction error quantifies the difference between the predicted and
the actual outcome, reflecting the choice and whether it was
subsequently rewarded:

\begin{equation}\phantomsection\label{eq-rl-pe}{
\delta_{t-1} = R_{t-1} - V_{t-1}^{\text{Work}}
}\end{equation}

\emph{Fictitious play.} The fictitious play model assumes that the
participant instead monitors their opponent's probability of selecting
an action (specifically, the probability of choosing ``Not Inspect'')
based on the opponent's previous play history:

\begin{equation}\phantomsection\label{eq-fp-belief}{
P_{t}^{\text{Not Inspect}} = P_{t-1}^{\text{Not Inspect}} + \eta \cdot \delta_{t-1}^{P}
}\end{equation}

This first-order belief is subsequently used to select the action that
maximises the expected outcome. The prediction error
\(\delta_{t-1}^{P}\) is updated:

\begin{equation}\phantomsection\label{eq-fp-pe}{
\delta_{t-1}^{P} = P_{t-1} - P_{t-1}^{\text{Not Inspect}}
}\end{equation}

Here \(P_{t-1}\) represents the observed action taken by the opponent; 1
if the employer chose ``Not Inspect'' and 0 if the opponent chose
``Inspect''. The participant's choice value in the model is subsequently
determined based on the payoff matrix:

\begin{equation}\phantomsection\label{eq-fp-value}{
V_{t}^{\text{Work}} = 50 - 100 \cdot P_{t}^{\text{Not Inspect}}
}\end{equation}

\emph{Influence learning.} The influence learning model integrates both
first- and second-order beliefs, suggesting that participants' decisions
are jointly influenced by the opponent's past choices and expectations
based on observing the participant's past choices:

\begin{equation}\phantomsection\label{eq-inf-belief}{
P_{t}^{\text{Not Inspect}} = P_{t-1}^{\text{Not Inspect}} + \eta \cdot \delta_{t-1}^{P} + \kappa \cdot \delta_{t-1}^{Q}
}\end{equation}

The second-order belief in the influence learning model is weighted by
κ, and the prediction error of the inferred opponent's belief about the
participant's action is calculated as:

\begin{equation}\phantomsection\label{eq-inf-pe2}{
\delta_{t-1}^{Q} = Q_{t-1} - Q_{t-1}^{\text{Work}}
}\end{equation}

where \(Q_{t-1}\) is the participant's action at the trial, recorded as
1 when the participant chooses ``Work'' and 0 when the participant
chooses ``Shirk''. \(Q_{t-1}^{\text{Work}}\) is the opponent's inferred
participant's working probability:

\begin{equation}\phantomsection\label{eq-inf-qwork}{
Q_{t-1}^{\text{Work}} = \frac{1}{5} - \frac{1}{5\beta} \cdot \log\left( \frac{1 - P_{t-1}^{\text{Not Inspect}}}{P_{t-1}^{\text{Not Inspect}}} \right)
}\end{equation}

\(P_{t}^{\text{Not Inspect}}\) is derived by combining the first-order
belief update \(\eta \cdot \delta_{t-1}^{P}\) and the second-order
belief update \(\kappa \cdot \delta_{t-1}^{Q}\), and is then used to
update the choice value \(V_{t}^{\text{Work}}\). Importantly, if the
weight κ for the second-order belief is equal to 0, the model reduces to
the fictitious play model.

\emph{Mixed-equilibrium reference.} The mixed-equilibrium (ME) model
{[}\citeproc{ref-hill2017}{106}{]} provides a non-learning baseline.
Each participant's choices are modelled as a Bernoulli process with a
single rate \(\delta_{s}\) (the probability of working), with no
dependence on trial history:

\begin{equation}\phantomsection\label{eq-me}{
Q_{t}^{s} \sim \text{Bernoulli}(\delta_{s})
}\end{equation}

Because the model is memoryless, its likelihood is scored on the same
observation set as the learning models (see below), so that the LOO
comparison against ME is valid.

\paragraph{Predictive fit: model fitting and
comparison}\label{predictive-fit-model-fitting-and-comparison}

All models were implemented in JAGS (v4.3.2)
{[}\citeproc{ref-plummer2003}{202}{]} and fitted via the \{rjags\} and
\{runjags\} interfaces in R v4.4.1 (2024-06-14). Predictive model
comparison was conducted using approximate leave-one-out
cross-validation (LOO-CV), with the pointwise log-likelihood evaluated
in R and the LOO Information Criterion (LOO-IC) computed via the \{loo\}
v2.7.0 package {[}\citeproc{ref-vehtari2017a}{203}{]}.

We used a hierarchical Bayesian structure for all groups, in which each
participant's parameters were drawn from a group-level distribution
described by a mean and a between-subject variance. The human group
represents a heterogeneous sample, and its group-level distribution was
estimated under weakly-informative priors. However, the LLM groups
comprise near-identical agents running the same model under a fixed
prompt. LLM and human models differed only in the group-level mean
hyperprior, for which we placed a weakly-informative logit-normal prior
(median 0.5) in place of the human Beta(1, 1). The between-subject
variance, freely estimated under the same prior as the human group,
returned near-zero values for every LLM group. An initial attempt to fit
the LLM groups under the same diffuse priors as the human group produced
unstable group-level estimates, consistent with this homogeneity.

\emph{Parameterisation.} The learning-rate parameters were bounded to
the unit interval. The first-order learning rate (\(\eta\)), the
second-order weight (\(\kappa\)) and the inverse temperature (\(\beta\))
were each drawn from a Beta distribution truncated to
\([0.001, 0.999]\). For the human group, the group-level Beta
distributions were parameterised by a mean and precision, with the mean
drawn from \(\text{Beta}(1,1)\) and the precision from
\(\text{Gamma}(1, 0.1)\). For the LLM groups, the same beta
mean/precision structure was used, but the group-level mean was given a
weakly-informative logit-normal prior (median 0.5) with the precision
again from \(\text{Gamma}(1, 0.1)\). The ME model used the same Beta
mean/precision form for its per-subject rate, with the precision drawn
from \(\text{Uniform}(0.001, 100)\).

\emph{Decision rule.} For the influence and fictitious play models, the
belief \(q_{t}\) that the employer would not inspect was updated
trial-by-trial and mapped to a probability of working via a logistic
decision rule:

\begin{equation}\phantomsection\label{eq-dr-belief}{
P(\text{Work})_{t} = \sigma\!\left(15\beta\,(2 - 4q_{t})\right)
}\end{equation}

where the boundary-separation coefficient is pinned at 15. For the RL
model, the probability of working was derived from the difference in
action values:

\begin{equation}\phantomsection\label{eq-dr-rl}{
P(\text{Work})_{t} = \sigma\!\left(15\beta\,(V_{t}^{\text{Work}} - V_{t}^{\text{Shirk}})\right)
}\end{equation}

\emph{Convergence and comparison.} All models were estimated using 3
independent Markov chains. Convergence was assessed using the
Gelman-Rubin statistic (\(\widehat{R}\))
{[}\citeproc{ref-gelman1992}{204}{]} and the effective sample size
(ESS), with \(\widehat{R}\) \textless{} 1.01 and ESS \textgreater{} 400
for all reported parameters. Models were compared within each group
using LOO-CV, which approximates the expected log pointwise predictive
density (ELPD) for new data {[}\citeproc{ref-vehtari2017a}{203}{]},
where lower LOO-IC values indicate a better out-of-sample predictive
accuracy.

\paragraph{Model--behaviour fit: parameter--behaviour
correspondence}\label{modelbehaviour-fit-parameterbehaviour-correspondence}

To assess the construct validity of the fitted parameters we regressed
three model-free behavioural summaries on the fitted η and κ. Parameters
were taken from the influence-play fit for every group where it
presented as the winning model. The predictors were the fitted η and κ,
which were log-transformed and then z-scored across the pooled analysis
set, so that coefficients are comparable across groups.

For each group we fitted three separate ordinary-least-squares (OLS)
linear regressions using the \texttt{lm()} function from the R package
\{stats\} v4.4.1, one per behavioural outcome, each entering η and κ as
predictors:

\begin{itemize}
\tightlist
\item
  Trial-to-trial switching:
  \(p(\text{Switch})_i \sim \eta_i + \kappa_i\)
\item
  Lagged tracking of the employer's previous move:
  \(r(\text{employer action}_{t-1}, \text{choice}_t)_i \sim \eta_i + \kappa_i\)
\item
  Task performance: \(\text{mean payoff}_i \sim \eta_i + \kappa_i\)
\end{itemize}

As DeepSeek-SCoT's winning model was fictitious play, we ran the same
three regressions for DeepSeek-SCoT using its own winning
fictitious-play fit, with η as the sole predictor. Because this places
DeepSeek-SCoT's η on a different z-scale from the pooled set used for
the other groups, its coefficients are not directly comparable and are
marked in Table S3.

Coefficients are reported as standardised slopes with their standard
errors and two-sided p-values. Given the substantial differences in
parameter variance, we additionally computed the scale-free partial
correlation which expresses each effect within the group on a bounded
{[}−1, 1{]} scale.

\paragraph{Generative fit: posterior predictive
check}\label{generative-fit-posterior-predictive-check}

For the posterior predictive check, we drew 2,000 samples from each
group's winning-model posterior, using each draw's per-participant
parameter values to simulate one full replicate dataset - the same
number of participants and trials as observed - by running the model
forward. For each statistic - the overall probability of working, the
switch rate and the mean payoff - we computed the group-level mean of
that statistic across all simulated participants, yielding a 2,000-value
distribution of simulated group means. Subtracting the single fixed
observed group mean from every draw gave a discrepancy distribution
which centred on zero when the model reproduces the observed level and
is shifted away from zero when it does not.

We examined each statistic with the region-of-practical-equivalence
(ROPE) equivalence test {[}\citeproc{ref-kruschke2018}{119}{]}, in the
full-posterior form following recommended guidelines
{[}\citeproc{ref-makowski2019}{120}{]}. Subsequently, we computed the
percentage of the distribution lying within a ROPE of ±0.05, and
accepted a statistic as practically reproduced when this exceeded
97.5\%, rejected it below 2.5\%, and left it undecided in between. The
ROPE half-width was fixed at 0.05 on the shared {[}0,1{]} scale on which
all three statistics lie. The percentage in ROPE per statistic is given
in Table S4.

\subsubsection{Rock-paper-scissors}\label{rock-paper-scissors-5}

\paragraph{Candidate models}\label{candidate-models-1}

Fitted models were identical to a previous study investigating adaptive
mentalization among human participants
{[}\citeproc{ref-buergi2026}{107}{]}, which we only briefly summarise
below. We refer readers to the original publication for a more
comprehensive description and formulation of the models.

\paragraph{Cognitive Hierarchy Assessment (CHASE)
model}\label{cognitive-hierarchy-assessment-chase-model}

The CHASE model works on the premise that there is an action that
non-strategic players (defined as k = 0) would choose, whereas strategic
players add a limited number of recursive reasoning steps by iteratively
best-responding to that action. Importantly, adaptive agents infer the
level of recursive reasoning of the opponent by integrating evidence for
the different levels over time.

More formally, non-strategic play (k = 0) is governed by an updating
rule which maps the history of the game to attractions for each action
at each trial. In this context, attractions are a noisy representation
of historical action frequencies with an exponential memory decay. These
attractions are linked to the observed behaviour, governing the extent
to which agents tend to stick to the historical action frequencies or
act randomly.

While a level-0 agent solely responds to their own action frequencies,
higher-level agents try to predict what the opponent will play, or
predicting what the opponent thinks they will play, etc. Specifically,
an agent with sophistication k assumes that the sophistication of the
other agent is exactly one level lower (i.e.~k - 1) and therefore
applies k steps of recursive reasoning. While a k = 1 agent is certain
that the opponent must be k = 0, higher-level agents face uncertainty
about which level the opponent is most likely playing. To adapt to
different strategic players, adaptive agents form and update beliefs
about the opponent's level of sophistication. Agents then form an
integrated prediction over the most likely opponent action, weighted by
the belief distribution over the opponent's level, and noisily best
respond to it.

In total, the resulting model is characterized by 5 parameters: α, β, γ,
λ and κ (regulating the speed of updating attractions, recursive
reasoning noise, sensitivity to evidence for opponent's level, loss
aversion, and depth of mentalization ability, respectively).

\paragraph{Alternative models}\label{alternative-models}

\emph{Reinforcement learning.} A simple non-strategic learning rule is
to repeat the actions that were successful in the past. Specifically,
the model updates a value for each action at a learning rate α and
selects actions via a softmax over these values with inverse temperature
β, learning only from its own payoffs with no representation of the
opponent.

\emph{Fictitious play.} A more sophisticated approach than simple
reinforcement learning is fictitious play, where agents try to estimate
the probabilities with which the other player chooses her actions, and
then best respond to them. This agent corresponds to a CHASE model with
κ = 1 (i.e.~equivalent to k = 1). Both fictitious play and reinforcement
learning are fully specified by two free parameters, a learning rate α
and a temperature parameter β.

\emph{Experience-weighted attraction (EWA).} EWA
{[}\citeproc{ref-camerer1999}{205}{]} is a hybrid model combining
elements of reinforcement learning and fictitious play. In brief, this
is achieved by introducing a parameter δ that quantifies the extent to
which an agent also learns from foregone payoffs (i.e.~the payoffs that
would have resulted when choosing different actions). Our formulation of
EWA has three free parameters (δ, ϕ, ρ) that govern the updating
process, where we manually set the initial values for attractions and
experience-equivalents that are free parameters in the original
formulation.

\emph{Self-tuning EWA.} This is a simplified version of the EWA
{[}\citeproc{ref-ho2007}{206}{]} that fixes some parameter values to
empirical values and replaces others with functions of experience. As a
result, only the behavioural temperature parameter β is estimated.

\emph{Theory-of-Mind k (ToMk).} Similar to CHASE, the ToMk model
{[}\citeproc{ref-weerd2018}{122}{]} also involves simulating what
opponents of increasing sophistication would play, but uses a heuristic
confidence-updating mechanism to form a response, rather than a Bayesian
update on which the CHASE model is based.

Specifically, the level-0 agent is defined as noisily responding to
recency-weighted past action frequencies of the opponent. The level-1
agent simulates the opponent's level-0 behavior, computes their most
likely action and then responds combining the best response to this
opponent prediction with the expected value from their `own' level-0
strategy. Finally, higher levels simulate this response process from the
opponent's perspective and then recursively integrate higher-level
responses with a confidence for each level.

The ToMk model consists three free parameters: a learning rate α
governing both the frequency and confidence updates, an inverse
temperature β controlling choice stochasticity, and an integer κ setting
the highest level of recursive reasoning the agent performs.

\paragraph{Model fitting and
comparison}\label{model-fitting-and-comparison}

To fit the models to our participants' behaviour, we used maximum
likelihood estimation by combining an initial grid-search with the
\emph{fminunc} function in MATLAB. We fitted one set of parameters per
subject (i.e.~across opponents) for each model, but reset all relevant
prior belief variables to uniform upon facing a new opponent where
appropriate (e.g.~beliefs, attractions, etc) (Fig S2). We computed
negative log-likelihood (negLL), Akiake Information Criterion (AIC) and
Bayesian information Criterion (BIC) scores for model comparison metrics
whilst accounting for differences in the number of free parameters. We
used random effects Bayesian model comparison to ensure that the
comparison is not overly sensitive to outliers, using the VBA toolbox
{[}\citeproc{ref-daunizeau2014}{207}{]}. We computed the winning model
through the protected exceedance probability (PXP), quantifying the
likelihood that a model is expressed more frequently than all other
candidates, accounting for the possibility of chance differences.

\newpage

\subsection*{Data and code
availability}\label{data-and-code-availability}
\addcontentsline{toc}{subsection}{Data and code availability}

Data and code for modeling and analysis are available on request.

\subsection*{Acknowledgements}\label{acknowledgements}
\addcontentsline{toc}{subsection}{Acknowledgements}

A.S. was funded by the MRC AIM and AIM iCASE Grant (MR/W007002/1). L.Z.
was supported by the Wellcome Trust (228268/Z/23/Z), the Royal Society
(IES\textbackslash R3\textbackslash243253), and the Birmingham--Nanjing
University Joint Research and Education Centre pump-priming funding. The
computations described in this paper were partially performed using the
Birmingham Environment for Academic Research (BEAR). P.L.L. was
supported by a Medical Research Council Fellowship (MR/P014097/1 and
MR/P014097/2), a Jacobs Foundation Research Fellowship, a Leverhulme
Prize (PLP-2021-196), a Wellcome Trust/Royal Society Sir Henry Dale
Fellowship (223264/Z/21/Z) and a UKRI EPSRC Frontiers Research
Guarantee/ERC Starting Grant (EP/X020215/1). We thank Ian Apperly,
Robert Lee, Ayat Abdurahman, Daniel Drew and Luca Hargitai for their
help with data collection, and Niklas Buergi for helpful discussions
regarding the modelling analyses.

\subsection*{Author contributions}\label{author-contributions}
\addcontentsline{toc}{subsection}{Author contributions}

A.S.: Conceptualization, Methodology, Software, Validation, Formal
Analysis, Investigation, Data Curation, Writing -- Original Draft,
Writing -- Review \& Editing, Visualization, Project Administration.\\
X.Z.: Investigation, Data Curation, Writing -- Review \& Editing\\
A.K.: Resources, Writing -- Review \& Editing, Supervision\\
P.L.L.: Resources, Writing -- Review \& Editing, Supervision, Funding
acquisition\\
L.Z.: Conceptualization, Methodology, Validation, Resources, Writing --
Review \& Editing, Supervision, Project administration, Funding
acquisition

\subsection*{Competing interests}\label{competing-interests}
\addcontentsline{toc}{subsection}{Competing interests}

The authors declare no competing interests.

\subsection*{Declaration of generative
AI}\label{declaration-of-generative-ai}
\addcontentsline{toc}{subsection}{Declaration of generative AI}

In the preparation of this manuscript, agentic workflows using Claude
Code (models Sonnet, Opus and Fable) were used to screen text supplied
by the authors, and for AI-assisted programming. However, all
programming outputs were verified and screened by the lead author, and
no original text was generated by the agents. Use of generative AI in
this manuscript adheres to ethical guidelines for use and acknowledgment
of generative AI in academic research
{[}\citeproc{ref-hosseini2025}{208},
\citeproc{ref-porsdammann2024}{209}{]}.

\newpage

\section*{References}\label{bibliography}
\addcontentsline{toc}{section}{References}

\phantomsection\label{refs}

\end{document}